\pdfoutput=1

\documentclass[11pt]{article}

\usepackage[final]{acl}

\usepackage{times}
\usepackage{latexsym}

\usepackage[T1]{fontenc}

\usepackage[utf8]{inputenc}

\usepackage{microtype}

\usepackage{inconsolata}
\usepackage{booktabs}
\usepackage{soul}
\usepackage{color}

\usepackage{multirow}
\usepackage{multicol}
\usepackage{enumitem,kantlipsum}
\usepackage[normalem]{ulem}
\usepackage{color,colortbl}
\usepackage{todonotes} 
\usepackage{soul}
\usepackage{threeparttable, makecell}
\usepackage{makecell}
\usepackage{xcolor}

\usepackage{tcolorbox}

\usepackage{comment}
\usepackage{array}  
\usepackage{lscape}  
\usepackage{listings} 
\usepackage{multirow}
\usepackage{graphicx}
\usepackage[normalem]{ulem}
\useunder{\uline}{\ul}{}

\usepackage{array}  
\usepackage{multirow}  
\usepackage{geometry}  
\usepackage{algorithm}
\usepackage{algpseudocode}
\usepackage{amsmath}
\usepackage{subcaption}
\usepackage{fancybox}
\usepackage{tikz}
  
\usepackage{graphicx}
\definecolor{lightblue}{rgb}{.50,.95,1}
\definecolor{tri}{rgb}{.25,.88,.82}
\definecolor{lilac}{rgb}{0.85,0.64,0.85}

\newcommand{\nqa}{\emph{NativQA}}
\newcommand{\mnqa}{Multi\emph{NativQA}}
\newcommand{\rebuttal}[1]{\textcolor{black}{#1}}
\usepackage{hyperref}
\usepackage{verbatim}
\usepackage{setspace}

\usepackage{colortbl}
\usepackage{booktabs}
\usepackage{arabtex}
\usepackage{utf8}
\setcode{utf8}

\title{NativQA: Multilingual Culturally-Aligned Natural Query for LLMs} 

\author{
    Md Arid Hasan,\textsuperscript{$^*$$\dagger$}\textsuperscript{$^1$}
    Maram Hasanain,\textsuperscript{$^2$}
    Fatema Ahmad,\textsuperscript{$^2$}
    Sahinur Rahman Laskar,\textsuperscript{$^3$}\\
    {\bf
    Sunaya Upadhyay,\textsuperscript{$^4$}
    Vrunda N Sukhadia,\thanks{The contribution was made while the author was interning at the Qatar Computing Research Institute.}\textsuperscript{$^2$}
    Mucahid Kutlu,\textsuperscript{$^5$}
    } \\
    {\bf
    Shammur Absar Chowdhury,\textsuperscript{$^2$}
    Firoj Alam\thanks{Equal contribution.}\textsuperscript{$^2$}
    } \\
    \textsuperscript{$^1$}University of New Brunswick, Canada, 
    \textsuperscript{$^2$}Qatar Computing Research Institute, Qatar, \\  
    \textsuperscript{$^3$}UPES, India, 
    \textsuperscript{$^4$}Carnegie Mellon University in Qatar, Qatar,
    \textsuperscript{$^5$}Qatar University, Qatar \\   
    arid.hasan@unb.ca, fialam@hbku.edu.qa \\
}

\begin{document}
\maketitle
\begin{abstract}
\end{list}
Natural Question Answering (QA) datasets play a crucial role in evaluating the capabilities of large language models (LLMs), ensuring their effectiveness in real-world applications. Despite the numerous QA datasets that have been developed and some work has been done in parallel, there is a notable lack of a framework and \textit{large scale region-specific datasets} queried by native users in their own languages. This gap hinders the effective benchmarking and the development of fine-tuned models for regional and cultural specificities. In this study, we propose a scalable, language-independent framework, \nqa, to seamlessly construct culturally and regionally aligned QA datasets in native languages, for LLM evaluation and tuning. We demonstrate the efficacy of the proposed framework by designing a multilingual natural QA dataset, \mnqa, consisting of $\sim$64k manually annotated QA pairs in seven languages, ranging from high to extremely low resource, based on queries from native speakers from 9 regions covering 18 topics. We benchmark open- and closed-source LLMs with the \mnqa{} dataset. We made the 
\mnqa{} dataset,\footnote{\href{https://huggingface.co/datasets/QCRI/MultiNativQA}{https://huggingface.co/datasets/QCRI/MultiNativQA}} and other experimental scripts\footnote{\href{https://gitlab.com/nativqa/multinativqa}{https://gitlab.com/nativqa/multinativqa}} publicly available for the community. 
\end{abstract}

\section{Introduction}
\label{sec:introduction}



\begin{figure}[t]
    \centering    
    \includegraphics[scale=0.35]{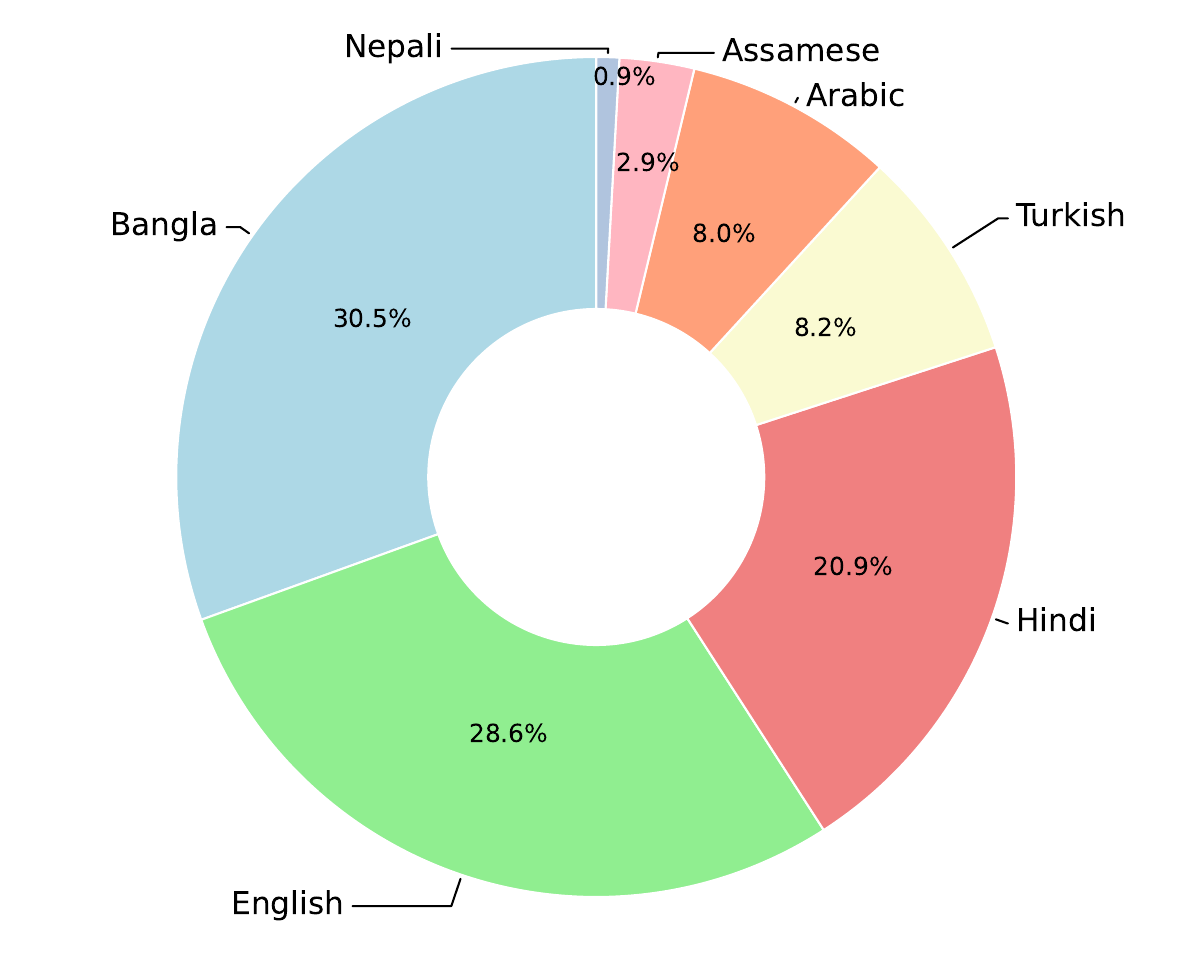}
    \vspace{-0.2cm}
    \caption{Distribution of the \mnqa{} dataset across different languages.}
    \label{fig:nativqa_dataset_plot}
    \vspace{-0.4cm}
\end{figure}

Recent advancements in LLMs have revolutionized the landscape of artificial intelligence, significantly pushing the state-of-the-art for a broad array of Natural Language Processing (NLP) and Speech Processing tasks. 
Their potential in language understanding and generation, across multiple (high- and low-resourced) languages, has attracted researchers to integrate and benchmark the LLM capabilities across diverse tasks, domains, and disciplines~\citep{openai2023gpt4,touvron2023llama}. 
However, the rapid integration of LLMs necessitates measuring cultural discrepancies in the responses generated by LLMs to ensure alignment with users' cultural values and contexts~\citep{naous-etal-2024-beer,alkhamissi-etal-2024-investigating,shen-etal-2024-understanding,liu-etal-2024-multilingual,arora2024calmqa,myung2024blend}.
This is particularly crucial in cross-lingual scenarios, where LLMs hallucinate or produce stereotypical responses biased toward Western culture, neglecting diverse cultural norms~\citep{naous-etal-2024-beer}. Consequently, such biases hinder the effectiveness of LLMs in daily-use applications for diverse languages and cultures, largely due to their under-representation in the training data used for these models.

\begin{figure*}[!ht]
    \centering    
    \includegraphics[scale=0.9]{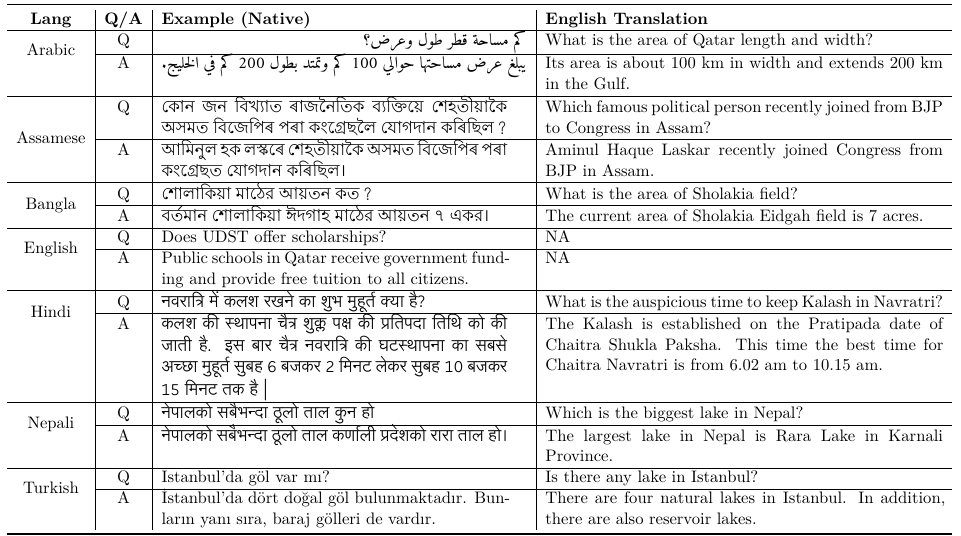}        
    \caption{Examples of questions and answers in different languages with their translation from our dataset.}
    \vspace{-0.3cm}
    \label{fig:examples_qa}
\end{figure*}

There are limited 
multilingual region-specific cultural benchmarks designed to evaluate the LLMs' performance across different cultures and languages. As a result, multilingual and non-English LLMs have been evaluated by using MT, with or without human involvement, to translate the existing English datasets into corresponding languages \citep{fanar2024}. 
However, translation often misses the cultural and regional nuances of target languages, making human-annotated datasets a better alternative.
In a recent study, ~\citet{arora2024calmqa} developed 1.5K culture-specific QAs by gathering questions from community web forums and employing native speakers to manually write questions. Similarly, \citet{myung2024blend} produced 52.5K multiple-choice and short-answer questions, with both question collection and answer writing being fully manual. 


\begin{figure}[h]
    \centering    
    \frame{\includegraphics[scale=0.62]{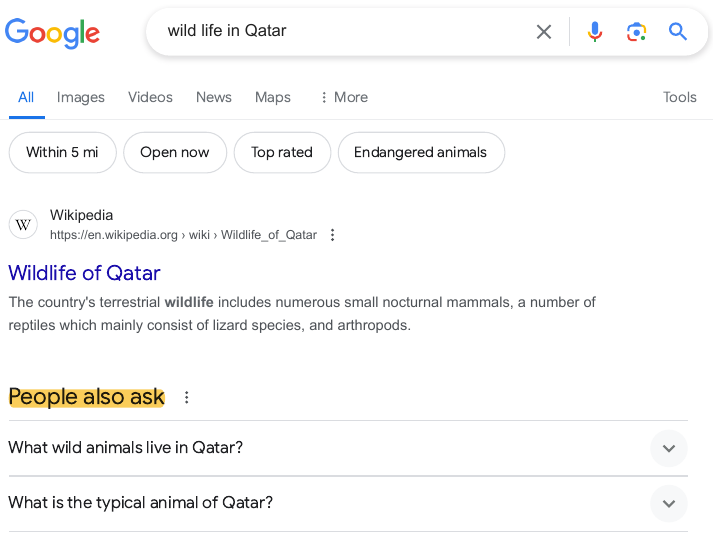}}        
    \caption{Google's QA list in response to a query.}
    \label{fig:google_example}
    \vspace{-0.3cm}
\end{figure}



In this study, we propose a framework, \textbf{Nativ}e \textbf{QA} 
(\nqa{}), specifically designed to seamlessly develop regionally- and culturally- specific QA datasets following a human-machine collaborative approach. Datasets developed through \nqa{} serve two primary functions: \textit{(i)} evaluating the LLM performance over real users' information needs and interests expressed in their native languages, and \textit{(ii)} facilitating fine-tuning of LLMs to adapt to cultural contexts. Moreover, to show the efficacy of the \nqa{} framework, we developed a natural \textbf{Multi}lingual  \textbf{Nativ}e question-answering (\textbf{QA}) dataset, \mnqa{}, including $\sim64k$ QA pairs in seven extremely low to high resource languages (see in Figure \ref{fig:nativqa_dataset_plot}), covering $18$ different topics from nine different regions (see examples in Figure \ref{fig:examples_qa}).
We further demonstrate the usefulness of \mnqa{} dataset by fine-tuning Llama-3.1. 


Unlike ~\citet{arora2024calmqa, myung2024blend}, the proposed \nqa{} framework can seamlessly collect QA pairs with minimal human intervention. Additionally, the answers are grounded in web-based reference sources. Our approach is inspired by the regional-based search engine queries addressing everyday needs as shown in Figure \ref{fig:google_example}.
Below we provide \textbf{our contributions} of this study: 
\begin{itemize}[noitemsep,topsep=0pt,leftmargin=*,labelsep=.5em]
    \item We propose the semi-automatic -- \nqa{} framework for developing culture- and region-specific natural QA datasets, enhancing LLMs inclusivity and providing comprehensive, culturally aligned benchmarks.
    \item We develop and release the \mnqa{} dataset, in seven languages with $\sim64k$ manually annotated QA pairs, covering $18$ different topics from native speakers across nine different regions. Additionally, we have collected another $55k$ QA pairs from six different locations \rebuttal{developed using our semi-supervised approach.}
   \item We benchmark over \mnqa{}~ with $2$ open and $2$ closed LLMs. In addition, we report experimental results of a fine-tuned Llama-3.1 model across all languages.
\end{itemize}
A summary of our findings is as follows:

\noindent
\textbf{Gap -- High vs. Low Resources Languages.} 
We observed the highest performance for English and lowest for Assamese on average across models, which clearly indicates that the performance correlates to the representation and/or richness of digital content of the language used in the models. This finding corroborates the findings reported in several parallel works \cite{myung2024blend}. 

\noindent
\textbf{Gap in Close vs. Open Models.} 
Close models outperforms open models. GPT-4o (BLEU: 0.230) and Gemini (BLEU: 0.226) perform similarly among closed models. Among open models, Llama-3.1 (BLEU: 0.186) outperforms Mistral (BLEU: 0.162).

\noindent
\textbf{Capability Enhancement with Fine-tuning.} 
Fine-tuning \textit{(i)} improves performance for extremely low resource languages such as Assamese and Nepali, \textit{(ii)} for medium resource languages, it helps dialect-rich languages like Arabic compared to other medium resource-languages (e.g., Hindi).

\noindent
\textbf{Cultural Benchmarking.}
Our findings emphasize the importance of well-crafted benchmarks efforts for studying regional/cultural awareness in LLMs. The results support the hypothesis that under-represented regions, and dialectal-rich languages (e.g., Arabic) benefit more from incorporating native and culturally aware information in the LLM. This highlights the value of the proposed language-independent framework \nqa{}, which efficiently creates multilingual, region- and culture-specific resources with minimal human effort.

\begin{figure*}[!htb]
    \centering    
\includegraphics[scale=0.3]{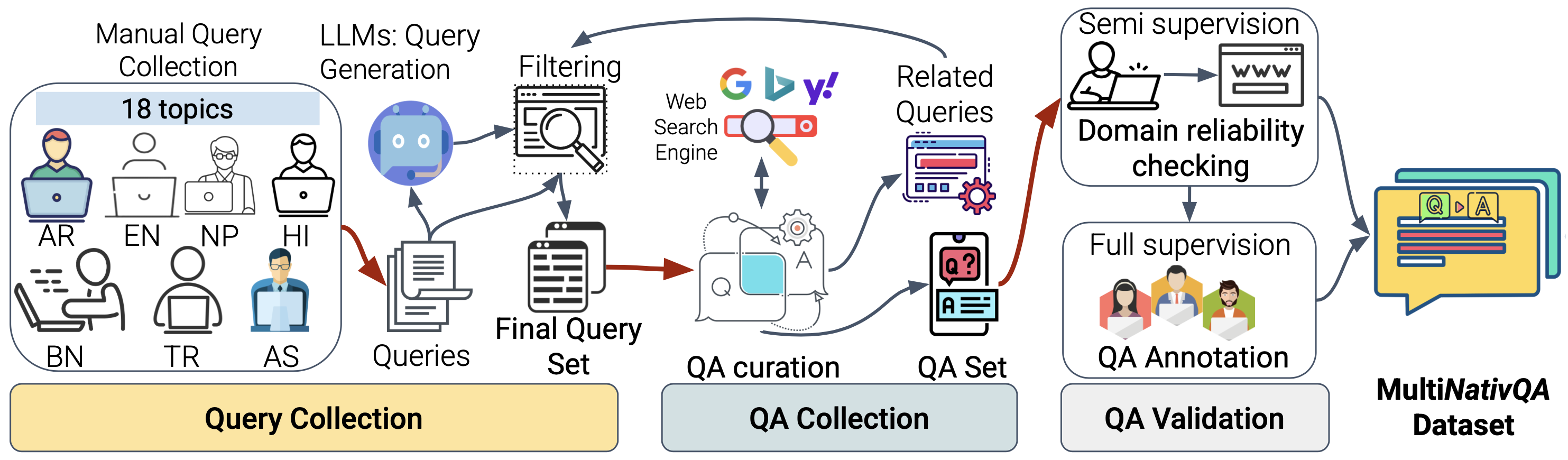}
    \vspace{-0.2cm}
    \caption{\nqa{} framework, demonstrating the data collection and annotation process. The details of each component of \nqa{} framework are discussed in Section \ref{sec:nativqa_framework}.}
    \label{fig:nativqa_pipeline}
    \vspace{-0.3cm}
\end{figure*}

\section{Related Work}
\label{sec:related work}

LLMs have demonstrated remarkable capabilities across various disciplines and tasks, leading to efforts to evaluate their performance on standard NLP tasks \citep{bubeck2023sparks,bang2023multitask,ahuja2023mega,hendy2023good}. While several initiatives have developed resources to benchmark LLMs, most focus primarily on English. For other languages, evaluations often rely on translated data \cite{lai2023okapi,sengupta2023jais,huang2024acegpt}.

\noindent\textbf{Existing QA Datasets.}
Question Answering has been a standard NLP task for decades, pushing the development of many QA datasets in different languages. 
\citet{kwiatkowski-etal-2019-natural} and \citet{yang-etal-2018-hotpotqa} proposed two extractive QA datasets including Natural Questions (NQ), both containing long-form 
large-scale question-answer pairs.
\citet{joshi-etal-2017-triviaqa} developed TriviaQA dataset, which consists of 650k question-answer-evidence triples. These triples are created by merging 95k question-answer pairs. \citet{rajpurkar-etal-2016-squad} developed SquAD, which is a collection of 100k crowdsourced 
QA's paired with shortened Wikipedia articles. 
HelpSteer \citep{wang2023helpsteer} is another QA dataset, which comprises a 37k sample dataset with multiple attributes of helpfulness preference. 
The most closest work in the literature to ours is BLEnD  \cite{myung2024blend} which is a hand-crafted benchmark consisting of 52.6k multiple choice and short-answer QA pairs for 13 different languages in total, focusing cultural aspects of languages.  

\noindent\textbf{Evaluations of LLMs for QA.}
For LLM evaluation, 
there are notable datasets covering world knowledge \citep{hendrycks2020measuring}, commonsense reasoning \citep{zellers2019hellaswag}, reading comprehension \citep{bandarkar2023belebele}, factuality \citep{lin2022truthfulqa}, and others. These datasets are usually transformed into multiple-choice questions. In addition, standard QA datasets have also been used for LLM evaluation \citep{hu2020xtreme}. \citet{kamalloo-etal-2023-evaluating} performed the analysis of different open-domain QA models, including LLMs by manually judging answers on a benchmark dataset of NQ-open \citep{lee-etal-2019-latent}.
Their investigation shows that LLMs attain state-of-the-art performance but fail in lexical matching when candidate answers become longer. 
In Table \ref{tab:existing_data} (Appendix), we report the most notable existing QA datasets compared to ours. 
\textbf{Compared to existing datasets}, the \mnqa{} dataset is novel in its topical coverage, with a focus on cultural aspects and regional nativeness. Furthermore, most recent cultural datasets are primarily designed for benchmarking purposes, whereas we also focused on model training.



\section{NativQA Framework}
\label{sec:nativqa_framework}

Figure \ref{fig:nativqa_pipeline} presents the \nqa{} framework with three inter-connected modules described below.



\subsection{Query Collection (QC)}
\label{ssec:qc}
The objective of this module is to collect open-ended queries, $\varrho$, centered on various predetermined topics derived from common concepts in everyday communication.
The topic set is first manually constructed. This manual effort allows us to identify topics that are culture- or region-specific. Examples of seed topics include: \textit{Animals, Business, Clothing, Education, Events, Food \& Drinks, General, Geography, Immigration, Language, Literature, Names \& Persons, Plants, Religion, Sports \& Games, Tradition, Travel}, and \textit{Weather}. 
However, \nqa{} framework is designed to be extensible and adaptable to any topic and language, not limited to 
high-coverage world knowledge.

Following, we start collecting the manual query set $\varrho_m$. We began by recruiting native speakers of the language of the target countries. Each speaker is encouraged to write $m$ queries per topic, in their native or second language,\footnote{Widely used in the respective city.} focusing on queries they might ask a search engine as residents of a corresponding major city. 
We then generate synthesized queries, $\varrho_s$, using initial seed queries, $\varrho_m$, and expand the $\varrho_m$ set with $\varrho_s$.
Synthesizing queries helps to increase the diversity in sub-topics and improve the versatility of writing styles in the final set of queries. It also reduces the skewness of the seed queries. For $\varrho_s$, we prompted an LLM to generate $x$ similar queries for each input query, $\varrho_{m}^{i} \in \varrho_m$. Finally, $\varrho_s$ is de-duplicated against $\varrho_m$ using exact string matching, resulting in the \textit{final set} of seed queries, $\varrho_0 = \varrho_m \bigcup \varrho_s$.

\subsection{QA Collection (QAC)}

Next, leveraging a search engine, we automatically collect QA pairs that potentially cover queries $\varrho_0$. The \nqa{} framework features with three major search engines (i.e., Google, Bing, and Yahoo), however, for the \mnqa{} we used 
`Google' and capitalized its feature 
-- ``People also ask'', where it lists several questions, searched by real users and are potentially relevant to the initial user query, as shown in Figure \ref{fig:google_example}. Moreover, these questions $Q$ are associated with answers $A$ extracted by the search engine, along with the attribution, $L$ -- links to the sources of the answers. \rebuttal{Each search engine has location and language features, which we utilize to collect native and location-specific QA pairs.}

Our QA curation module implements Algorithm~\ref{alg:collect_qa_pairs}, using the seed queries $\varrho_0$ along with the number of iteration, $N_{iter}$, as input. For each iteration $i \in N_{iter}$, we collect QA pairs $P_{QA}^i$, and related queries $S\varrho^i_{rel}$ for each query, $q \in S\varrho$, and then pass it to the filtering module and update the current query set $S\varrho$. We repeat the process for all iterations to obtain the final QA set, $S_{QA}$ with enriched queries $S\varrho$.

\subsection{QA Validation (QAV)}
Following, we validate the extracted QA pairs, considering at least two aspects: \textit{(i)} the quality and answerability of questions, and \textit{(ii)} reliability and completeness of answers. 
We validate the QA pairs through the following steps.

\paragraph{Domain Reliability Check (DRC).} 
First, we extract a unique set of web-domains using the attribution\footnote{Answer-source links} $L$ from the extracted QA pairs, $S_{QA}$. We then manually classify each domain's reliability based on an annotation guideline specifically designed for this task, inspired by several relevant studies~\citep{selejan2016credibility,flanagin2007role,metzger2015psychological}.
Next, we filtered out the QA pairs to retain answers only from annotated reliable sources as we hypothesize that answers from web pages on reliable domains are likely to be trustworthy. 
We adopted this approach for its scalability and reduced manual effort in obtaining reliable QA pairs. The final domain list (e.g., BBC, Guardian) can further aid QA extraction for multiple languages, especially for fine-tuning data.

\noindent \paragraph{QA Annotation (QAA).} 
Although some domains are considered reliable, the content they host may not always be trustworthy due to unreliable user-generated content. To address this, we further refined our framework by manually checking and editing the curated QA pairs from reliable sources. 
For each QA pair, we apply four types of annotations. \textit{(i)} \textit{Question validation}: Human annotators verify questions' quality by classifying each question as ``Good question'' or ``Bad question''. 
We then proceed to the subsequent steps using only the questions classified as ``Good''.
\textit{(ii)} \textit{Question's relavancy to the location}: Annotators are asked to classify whether the question is related to the specified location. \textit{(iii) Answer categorization}: Annotators 
examine each QA pair and assess whether the answer provides sufficient information to satisfy the question, and categorize the answers based on the correctness (see Sec. \ref{ssec:qa_annotation}).
\textit{(iii)} \textit{Answer editing}: If an answer is incomplete or incorrect, annotators must edit it using content from the source Web page. To maintain scope and reliability, we limit them to the provided source pages. Detailed annotation guidelines are in Appendix \ref{ssec:app_answer_categorization}.

\begin{algorithm}[!htp]
\footnotesize
\caption{\footnotesize Collecting QA pairs using seed queries $\varrho_0$. $P_{QA}^i$: QA pair, $S\varrho_{rel}^i$: related queries. ExtractQA(*) and ExtractRelatedQueries (*) are functions that return questions, $Q$-answers, $A$ pairs with attribution $L$, and related queries, respectively, which are obtained from the search engine for a given query, $q$. DeDuplication (*) removes any duplicate entries from the set to ensure uniqueness.}
\label{alg:collect_qa_pairs}
\begin{algorithmic}[1]
\State \textbf{Input:}
\State \hspace{\algorithmicindent} Seed queries: \( \varrho_0 = \{\hat{\varrho_1}, \hat{\varrho_2}, \ldots, \hat{\varrho_m}\} \)
\State \hspace{\algorithmicindent} Number of iterations: \( N_{iter} \)
\State \textbf{Output:}
\State \hspace{\algorithmicindent} Set of QA pairs: \( S_{QA} \)
\State \hspace{\algorithmicindent} Set of enriched queries: \( S\varrho \)
\State $S_{QA} \gets \emptyset$
\State $S\varrho \gets \varrho_0$
\For {$i$ from 1 to $N_{iter}$}
    \State $P_{QA}^i \gets \emptyset$
    \State $S\varrho_{rel}^i \gets \emptyset$
    \For {$q \in S\varrho$}
        \State $(Q^q, A^q, L^q) \gets \text{ExtractQA}(q)$
        \State $P_{QA}^i \gets P_{QA}^i \cup \{(q', a', l') \mid q' \in Q^q, a' \in A^q, l' \in L^q\}$
        \State $S\varrho_{rel}^i \gets S\varrho_{rel}^i \cup \text{ExtractRelatedQueries}(q)$
    \EndFor
    \State $P_{QA}^i \gets \text{DeDuplication}(P_{QA}^i)$
    \State $S_{QA} \gets S_{QA} \cup P_{QA}^i$
    \State $S\varrho \gets S\varrho \cup S\varrho_{rel}^i$    
\EndFor
\State \textbf{return} $S_{QA}, S\varrho$
\end{algorithmic}
\vspace{-0.1cm}
\end{algorithm}

\section{Multi\nqa{} Dataset}

We demonstrate the effectiveness and scalability of the \nqa{} framework by creating a large-scale, multilingual \mnqa{} dataset. The \mnqa{} dataset spans over seven languages --
from high- to extremely low-resource and nine different location/cities. 
Our choice of languages for \mnqa{} was guided primarily by the authors' native proficiency, which allowed for more accurate annotation and evaluation.
\mnqa{} captures linguistic diversity, by including several dialects for dialect-rich languages like Arabic.
We also added two linguistic variations of Bangla to reflect differences between speakers in Bangladesh and West Bengal, India. Furthermore, we included English queries from Dhaka and Doha, where English is often used as a second language. 




\subsection{\nqa{} Framework Adaptation} 

\noindent\textbf{Query Collection} For multilingual QC, we started with predetermined topics (see Section \ref{ssec:qc}) derived from common concepts in everyday lives of users (see in Appendix \ref{sec:app_seed_queries}). Next, we asked the residents and the native speakers to write $10$ to $50$ queries\footnote{Without a strict limit, some topics exceeded 50 queries.} per topic about their major cities and urban areas. 
We then used GPT-4 to generate $10$ similar queries based on each input query (see Tab. \ref{tab:prompts_sim_query_gen} for similar query generation prompt) and applied de-duplication on the seed queries. 
The number of queries per region is reported in Table \ref{tab:stats_queries}.

\noindent\textbf{QA Collection} Using \textit{QAC Module} we enriched queries and QA pairs for each language and its respective city. We ran our collection algorithm for 3-7 $N_{iter}$ per region based on the convergence rate. We collected $\sim154K$ QA pairs across all languages (see Table \ref{tab:stats_queries}:\#QA). 

\noindent\textbf{QA Validation} The \textit{QAV} is the final (and optional) phase of the \nqa{} framework. It includes two steps: domain reliability check (DRC) and QA annotation (QAA). These steps ensure high quality of the dataset and can be executed to the entire dataset or only test split, depending on the cost and time constraints. 
We applied both the DRC and QAA steps to all target languages and regions of \mnqa{} dataset to create a high-quality resource for the research community  (see Sec. \ref{ssec:maual_annot}).


\subsection{Manual Annotation}
\label{ssec:maual_annot}



We briefly discuss the manual annotation effort for QAV phase in \nqa{} framework for developing \mnqa{} dataset. For more detail instruction and analysis see Appendix \ref{sec:app_domain_reliablity}.

\subsubsection{Domain Reliability Check}
The objective for the domain reliability check is to verify the credibility of the source domain, which can be used to judge the factuality and reliability of answers sourced from that domain. We adopt the following definition of the credibility of the domain/website: ``A credible webpage is one whose information one can accept as the truth without needing to look elsewhere. If one can accept information on a page as true at face value, then the page is credible; if one needs to go elsewhere to check the validity of the information on the page, then it is less credible'' \citep{schwarz2011augmenting}. Annotators were tasked to review each web domain to determine its credibility and assign one of the following four reliability labels: \textit{(i)} very reliable, \textit{(ii)} partially reliable, \textit{(iii)} not sure, \textit{(iv)} completely unreliable. \rebuttal{We provide a detailed definition and guideline in Sec. \ref{sec:app_domain_reliablity} (in Appendix). For each language, 3 annotators manually checked 3,181 domains, and we identified 2,080 domains as very reliable and eliminated 1,101 domains, resulting in 65.38\% reliable and 34.62\% unreliable domains.}

\begin{table*}[!htb]
\centering
\setlength{\tabcolsep}{2pt} 
\scalebox{0.8}{%
\begin{tabular}{lllrr||rrrr}
\toprule
 & & & &  & \multicolumn{4}{c}{\textbf{\# Final Annotated QA}} \\
 \toprule
\textbf{Lang.} & \textbf{Cat.} & \textbf{City} & \textbf{\# of SQ} & \textbf{\# of QA} & \textbf{Train} & \textbf{Dev} & \textbf{Test} & \textbf{Total}  \\
\midrule
Arabic & M & Doha & 3,664 & 12,311 & 3,649 & 492 & 988 & 5,129  \\
Assamese & X & Assam & 900 & 21,009 & 1,131 & 157 & 545 & 1,833  \\
Bangla & L & Dhaka & 889 & 13,688 & 7,018 & 953 & 1,521 & 9,492  \\
Bangla & L & Kolkata & 900 & 13,378 & 6,891 & 930 & 2,146 & 9,967  \\
English & H & Dhaka & 1,339 & 17,744 & 4,761 & 656 & 1,113 & 6,530  \\
English & H & Doha & 3,414 & 25,621 & 8,212 & 1,164 & 2,322 & 11,698 \\
Hindi & M & Delhi & 1,184 & 16,328 & 9,288 & 1,286 & 2,745 & 13,319 \\
Nepali & L & Kathmandu & 1,222 & 11,503 & -- & -- & 561 & 561   \\
Turkish & M & Istanbul & 900 & 23,143 & 3,527 & 483 & 1,218 & 5,228 \\
\midrule
\textbf{Total} & & & \textbf{14,412} & \textbf{154,725} & \textbf{44,477} & \textbf{6,121} & \textbf{13,159} & \textbf{63,757}\\
\bottomrule
\end{tabular}
}
\caption{Statistics of our \mnqa{} dataset including languages with initial seed queries, the number of QA pairs collected per language from different locations and the final annotated QA pairs. Lang.: Language, SQ: Seed Query, Cat.: Categorization in terms of high (H), medium (M), low (L), and extremely low (X) as per \cite{lai2023chatgpt}, 
-- Only testing split due to limited dataset size.
}
\label{tab:stats_queries}
\vspace*{-0.4cm}
\end{table*}

\subsubsection{QA Annotation}
\label{ssec:qa_annotation}

This step of the QAV involves four types of annotations. Below, we discuss the brief guidelines for each annotation.

\begin{enumerate}[noitemsep,topsep=0pt,leftmargin=*,labelsep=.5em]
    \item \textbf{Question validation:}
The purpose of this task is to evaluate the quality of the questions. The annotators classified whether the questions are ``Good'' or ``Bad'' based on the criteria discussed below. The choice of the two types of questions was inspired by the NQ dataset~\citep{kwiatkowski-etal-2019-natural}. Depending on the annotation, the annotator's subsequent tasks vary. If a question is marked as `good', they proceed to the next task for the QA pair; otherwise, they skip further annotation and move on to the next QA pair.
\item \textbf{Question's relavancy to the location:} 
The purpose of this annotation was to check whether the question is related to the intended location. For example, ``\textit{Why do Emirati men wear white robes?}'' is a question related to UAE. 
\item \textbf{Answer categorization:} An answer can be categorized into one of these categories: \textit{(i)} correct, \textit{(ii)} partially correct, \textit{(iii)} incorrect, and \textit{(iv)} the answer can’t be found in the source page. Complete definition for each category is provided in Appendix \ref{ssec:app_answer_categorization}.
\item \textbf{Answer editing:} This step ensures the answer is correct, fully responds to the question, and is fluent and informative. If the answer is incorrect or incomplete, annotators must check the source page to extract content that completes the answer, if available.
\end{enumerate}

\subsection{Annotation Task Setup}
\label{ssec:annotation_setup}
The annotation team consisted of native speakers of the respective languages, with English as their second language.
The annotators had diverse educational backgrounds, ranging from undergraduate students to those holding PhD degrees. The team was trained and monitored by language specific 
expert annotators. To ensure quality, periodic checks of random annotation samples were conducted, and feedback was provided. Three annotators were assigned to the DRC task, and the final label is assigned based on majority voting.
For the QAA task, each QA pair was annotated by two annotators for the test set. In cases of disagreement, a third annotator reviewed and revised the annotations. For the training and dev set, each QA pair was annotated by one annotator.
These choices were made to maintain a balance between annotation quality, time, and cost. \rebuttal{For the annotation, we hired a third-party company that manages the payment process for the annotators, who are compensated at standard hourly rates based on their location. 
The annotation process took approximately $\sim$1400 hours.} We utilized in-house annotation platform 
for the tasks discussed in Appendix \ref{sec:app_annotation_platform}. 



\subsection{Annotation Agreement}

We evaluate the Inter-Annotator Agreement (IAA) of manual annotations using the Fleiss' Kappa coefficient ($\kappa$) for the domain reliability tasks.
The Kappa ($\kappa$) values across the languages ranges from 0.52 to 0.66 (except for English being 0.37), 
which correspond to fair to substantial agreement 
\citep{landis1977measurement}. 
Note that we selected the final label where the majority agreed, meaning that we have above 66\% agreement on the final label. 
For the \textit{QA annotation task} (answer editing), we first directly select only the questions where both annotators agree. For the disagreed cases, another annotator revises them; ultimately, we select based on the agreement of at least two annotators. 
For the answer editing, on average this matching is 66.04\% across languages. This is higher than BLEnD benchmark \cite{myung2024blend}, which reported an agreement score of 63.2\%. 
In addition we have computed Levenshtein distance to understand how much edits has been done. The average edits across all languages are relatively low ($0.17$) indicating 
minimal edits has been done on the answers. In Appendix \ref{sec:data_annotation_analysis}, we provide further details. 

\subsection{Statistics and Analysis}
\label{ssec:data_analysis}


Figure \ref{fig:nativqa_dataset_plot} reports the initial data distribution across languages, irrespective of the country they were collected from. English, Arabic, and Bangla are higher in proportion due to the fact that \textit{(i)} English consists of data collected from Qatar and Bangladesh, \textit{(ii)} Arabic consists of queries from different dialects, 
and \textit{(iii)} Bangla consists of data from Bangladesh and India. 
\rebuttal{The average length for question and answer are 6 and 35 words, respectively (See Tab. \ref{tab:qa_avg_length})}
As Table \ref{tab:stats_queries} shows, 
 our annotation process resulted in a decrease in QA set size by half (comparing initial QA set (column \textit{\#QA}) to final QA set (column \textit{F.QA})). 
 We also faced a significant drop for Assamese and Nepali. This drop is due to the fact that the search engine returned QA pairs in non-native languages (in these cases, either Hindi or English) rather than the native language. 
As part of our process, we filtered out QA pairs that are not in the target language. We identify the native language using a language detection tool\footnote{\href{http://fasttext.cc/docs/en/language-identification.html}{language detection tool}} and then manually revise them.
Our final \mnqa{} dataset covers a wide range of topics in all languages with similar distribution (see Appendix Figure \ref{fig:topic_wise_dist1} and \ref{fig:topic_wise_dist2}). To assess the efficacy of the \nqa{} framework, we additionally collected $55k$ QA pairs from $6$ different locations (see in Appendix \ref{sec:app_additional_data}. 



\begin{table*}[]
\centering
\setlength{\tabcolsep}{2pt} 
\scalebox{0.83}{%
\begin{tabular}{@{}l|rrr|rrr|rrr|rrr|rrr@{}}
\toprule
\multicolumn{1}{c}{\textbf{Model}} & \multicolumn{1}{|c}{\textbf{F1}} & \multicolumn{1}{c}{\textbf{BLEU}} & \multicolumn{1}{c}{\textbf{Rou.}} & \multicolumn{1}{|c}{\textbf{F1}} & \multicolumn{1}{c}{\textbf{BLEU}} & \multicolumn{1}{c}{\textbf{Rou.}} & \multicolumn{1}{|c}{\textbf{F1}} & \multicolumn{1}{c}{\textbf{BLEU}} & \multicolumn{1}{c}{\textbf{Rou.}} & \multicolumn{1}{|c}{\textbf{F1}} & \multicolumn{1}{c}{\textbf{BLEU}} & \multicolumn{1}{c}{\textbf{Rou.}} & \multicolumn{1}{|c}{\textbf{F1}} & \multicolumn{1}{c}{\textbf{BLEU}} & \multicolumn{1}{c}{\textbf{Rou.}} \\ \midrule
\multicolumn{1}{c}{\textbf{}} & \multicolumn{3}{c}{\cellcolor[HTML]{add8e6}\textbf{Arabic}} & \multicolumn{3}{c}{\cellcolor[HTML]{dda0dd}\textbf{Bangla-IN}} & \multicolumn{3}{c}{\cellcolor[HTML]{ffdb58}\textbf{English-BD}} & \multicolumn{3}{c}{\cellcolor[HTML]{ffff99}\textbf{Hindi}} & \multicolumn{3}{c}{\cellcolor[HTML]{deb887}\textbf{Turkish}} \\ \midrule
GPT-4o & 0.839 & \textbf{0.280} & \textbf{0.044} & 0.821 & 0.226 & 0.009 & \textbf{0.651} & \textbf{0.384} & \textbf{0.284} & \textbf{0.865} & \textbf{0.296} & \textbf{0.050} & 0.768 & \textbf{0.226} & \textbf{0.252} \\
Gemini-1.5 & \textbf{0.840} & 0.228 & 0.038 & \textbf{0.833} & \textbf{0.251} & \textbf{0.014} & 0.631 & 0.259 & 0.251 & 0.800 & 0.171 & 0.036 & \textbf{0.773} & 0.164 & 0.229 \\ 
Llama-3.1  & \underline{0.528} & \underline{0.202} & \underline{0.037} & \underline{0.453} & \underline{0.132} & \underline{0.007} & \underline{0.636} & 0.280 & \underline{0.256} & \underline{0.604} & \underline{0.260}& \underline{0.035} & \underline{0.616} & \underline{0.217} & \underline{0.202} \\
Mistral & 0.487 & 0.148 & 0.034 & 0.418 & 0.108 & 0.005 & 0.620 & \underline{0.345} & 0.251 & 0.553 & 0.177 & 0.030 & 0.563 & 0.193 & 0.161 \\
 \midrule
\multicolumn{1}{c}{\textbf{}} & \multicolumn{3}{c}{\cellcolor[HTML]{90ee90}\textbf{Assamese}} & \multicolumn{3}{c}{\cellcolor[HTML]{f08080}\textbf{Bangla-BD}} & \multicolumn{3}{c}{\cellcolor[HTML]{ffb6c1}\textbf{English-QA}} & \multicolumn{3}{c}{\cellcolor[HTML]{afeeee}\textbf{Nepali}} & \multicolumn{3}{c}{\textbf{Avg.}} \\  \midrule
GPT-4o & 0.745 & 0.107 & \textbf{0.021} & 0.826 & 0.154 & 0.007 & \textbf{0.628} & 0.314 & \textbf{0.260} & \textbf{0.873} & 0.086 & 0.003 & 0.779 & \textbf{0.230} & \textbf{0.103} \\
Gemini-1.5 & \textbf{0.808} & \textbf{0.150} & 0.016 & \textbf{0.844} & \textbf{0.292} & \textbf{0.010} & 0.620 & 0.274 & 0.241 & \textbf{0.873} & \textbf{0.244} & \textbf{0.005} & \textbf{0.780} & 0.226 & 0.093 \\
Llama-3.1  & \underline{0.523} & \underline{0.029} & \underline{0.005} & \underline{0.840} & \underline{0.119} & \underline{0.005} & \underline{0.622} & 0.294 & \underline{0.247} & \underline{0.582} & \underline{0.138} & \underline{0.002} & \underline{0.600} & \underline{0.186} & \underline{0.088} \\
Mistral & 0.485 & 0.020 & 0.003 & 0.820 & 0.080 & 0.005 & 0.608 & \textbf{\underline{0.332}} & 0.236 & 0.504 & 0.056 & 0.002 & 0.562 & 0.162 & 0.081\\
 \bottomrule
\end{tabular}
}
\caption{Performance of different LLMs across languages. F1: F1 BERTScore, Rou.: Rouge1, Llama-3.1: Llama-3.1-8B-Instruct, Gemini-1.5: Gemini-1.5 Flash, Mistral: Mistral-7B-Instruct-v0.1. \textbf{Bold} results are best per column per language. \underline{Underlined} results are best across open models. \textbf{Avg} Average over languages.}
\label{tab:results}

\vspace{-0.4cm}
\end{table*}


\section{Experimental Setup}
\label{sec:experiments}

\noindent\textbf{Data Splits.} 
We split the data for each region into training (70\%), development (10\%), and test (20\%) sets using stratified sampling based on topics as labels. Given the small size of the Nepali data, we kept the full dataset for test purpose.
Annotations were done separately for each data split, with some data removed due to bad questions or incorrect answers. This resulted in inconsistencies in split proportions across languages (see Tab.~\ref{tab:stats_queries}).


\noindent\textbf{Models.}
We experiment with both open and close LLMs. For the close models we use GPT-4o~\citep{achiam2023gpt} and Gemini 1.5 Flash.\footnote{gemini-1.5-flash-preview-0514} For open models, we opt for 
Llama-3.1-8B-Instruct,\footnote{\href{https://huggingface.co/meta-llama/Llama-3.1-8B-Instruct}{Llama-3.1-8B-Instruct}} and Mistral-7B-Instruct-v0.1.\footnote{\href{https://huggingface.co/mistralai/Mistral-7B-Instruct-v0.1}{Mistral-7B-Instruct-v0.1}} We use zero-shot learning as our setup with all models.
For reproducibility, we use the same prompt, response format, output token limit, and decoding parameters (e.g., temperature set to 0) across all models. We designed the prompts using concise instructions, as reported in Appendix~\ref{sec:app:prompting_zshot}. All prompts and evaluation scripts are released as part of \textit{LLMeBench}~\cite{dalvi-etal-2024-llmebench}.\footnote{\href{https://llmebench.qcri.org/}{https://llmebench.qcri.org/}}

\noindent\textbf{Fine-tuning Models.}
We demonstrate the efficacy of \mnqa{} training split for all regions by finetuning an open LLM -- Llama-3.1-8B-Instruct model. To reduce the computational cost, we opt for PEFT using LoRA~\citep{hu2022lora}. We train the model in full precision (FP16). We use Adam optimizer, set the learning rate to $2e-4$, lora alpha to 16, lora $r$ to 64, maximum sequence length to 512, with a batch size of 16. We fine-tune the model for one epoch with no hyper-parameter tuning.


\noindent\textbf{Fine-tuning Instructions.} For fine-tuning, we create a diverse set of English instructions using template-based approach. We design the templates by prompting two close models: GPT-4o and Claude-3.5 Sonnet,\footnote{\href{https://www.anthropic.com/news/claude-3-5-sonnet}{claude-3-5-sonnet}} to generate 10 diverse instructions per model for the QA task for each language.
Following, during fine-tuning, we randomly select one from these templates and append to the QA pair to create the final instruction. During inference, we randomly select one instruction and use it to prompt both the base and the fine-tuned model. Examples of instructions and prompts are in Appendix~\ref{sec:app:prompting_inst}.
\noindent\textbf{Evaluation and Metrics.}
We evaluate model performance on the \mnqa{} test set using standard QA evaluation metrics. For lexical (n-gram) similarity, we employ BLEU and ROUGE, while for semantic similarity, we use the F1 score within BERTScore~\citep{zhangbertscore}. BERTScore is computed using contextual embeddings extracted from pre-trained BERT models. We leverage language-specific transformer models for embedding extraction (see Appendix, Table \ref{tab:language_models_bert_score}). In addition, we conduct LLM-as-a-judge and human evaluations.
For GPT-4o-as-a-judge, we use the pointwise LLM-as-judge approach with reference answers, as described in \cite{zheng2023judging}. Ratings are assigned on a scale from 1 to 10 (see Appendix \ref{sec:app_eval_llm-as-a-judge}). For human evaluation, we use a 5-point Likert scale to assess response accuracy and usefulness (see Appendix \ref{sec:app_subjective_evaluation}).




\begin{figure}[]
    \centering    
    \includegraphics[scale=0.3]{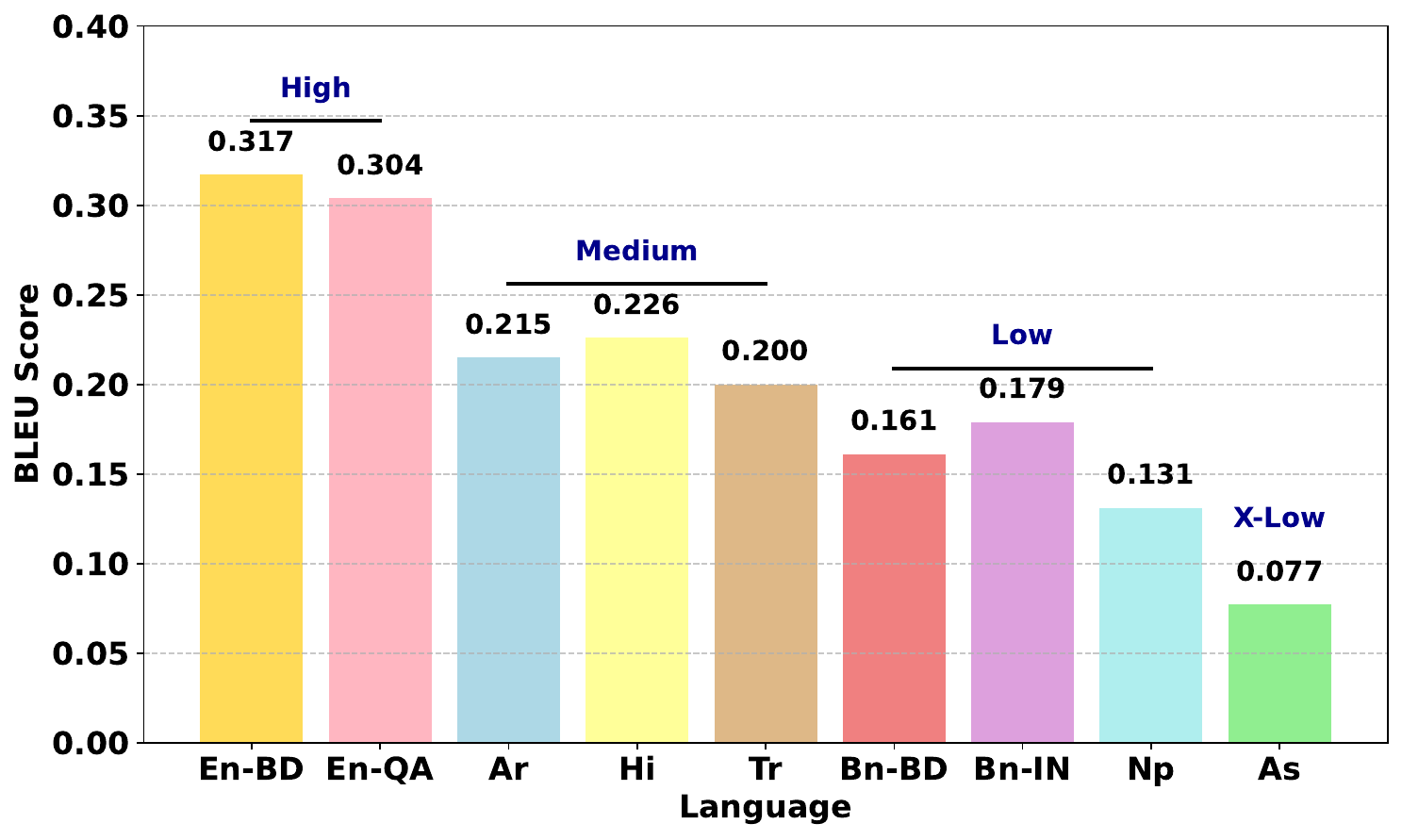}    
    \vspace{-0.4cm}
    \caption{Average performance (BLEU scores) of the models by language. X-Low: Extremely low.}
    \label{fig:bleu_score_plot}
    \vspace{-0.4cm}
\end{figure}

\begin{table*}[!htb]
\centering
\setlength{\tabcolsep}{2pt} 
\scalebox{0.81}{%
\begin{tabular}{@{}l|ccc|ccc|ccc|ccc|ccc@{}}
\toprule
\multicolumn{1}{c}{\textbf{Model}} & \multicolumn{1}{|c}{\textbf{F1}} & \multicolumn{1}{c}{\textbf{BLEU}} & \multicolumn{1}{c}{\textbf{Rou.}} & \multicolumn{1}{|c}{\textbf{F1}} & \multicolumn{1}{c}{\textbf{BLEU}} & \multicolumn{1}{c}{\textbf{Rou.}} & \multicolumn{1}{|c}{\textbf{F1}} & \multicolumn{1}{c}{\textbf{BLEU}} & \multicolumn{1}{c}{\textbf{Rou.}} & \multicolumn{1}{|c}{\textbf{F1}} & \multicolumn{1}{c}{\textbf{BLEU}} & \multicolumn{1}{c}{\textbf{Rou.}} & \multicolumn{1}{|c}{\textbf{F1}} & \multicolumn{1}{c}{\textbf{BLEU}} & \multicolumn{1}{c}{\textbf{Rou.}} \\ \midrule
\multicolumn{1}{c}{\textbf{}} & \multicolumn{3}{c}{\cellcolor[HTML]{add8e6}\textbf{Arabic}} & \multicolumn{3}{c}{\cellcolor[HTML]{dda0dd}\textbf{Bangla-IN}} & \multicolumn{3}{c}{\cellcolor[HTML]{ffdb58}\textbf{English-BD}} & \multicolumn{3}{c}{\cellcolor[HTML]{ffff99}\textbf{Hindi}} & \multicolumn{3}{c}{\cellcolor[HTML]{deb887}\textbf{Turkish}} \\ \midrule
Llama-3.1 & 0.508 & 0.080 & 0.032 & \textbf{0.451} & 0.054 & 0.005 & \textbf{0.621} & \textbf{0.247} & \textbf{0.234} & \textbf{0.606} & 0.123 & \textbf{0.038} & \textbf{0.613} & 0.092 & 0.188                \\
Llama-3.1-FT & \textbf{0.532} & \textbf{0.181} & \textbf{0.039} & 0.421 & \textbf{0.139} & \textbf{0.012} & 0.612 & 0.198 & 0.205 & 0.521 & \textbf{0.159} & 0.024 & 0.592 & \textbf{0.189} & \textbf{0.190}
\\ \midrule
\multicolumn{1}{c}{\textbf{}} & \multicolumn{3}{c}{\cellcolor[HTML]{90ee90}\textbf{Assamese}} & \multicolumn{3}{c}{\cellcolor[HTML]{f08080}\textbf{Bangla-BD}} & \multicolumn{3}{c}{\cellcolor[HTML]{ffb6c1}\textbf{English-QA}} & \multicolumn{3}{c}{\cellcolor[HTML]{afeeee}\textbf{Nepali}} & \multicolumn{3}{c}{\textbf{AVG}}  \\  \midrule
Llama-3.1 & 0.550 & 0.020 & 0.006 & \textbf{0.841} & 0.037 & 0.004 & \textbf{0.603} & \textbf{0.202} & \textbf{0.218} & \textbf{0.591} & 0.103 & 0.002 &\textbf{0.598} & 0.107 & \textbf{0.081}      \\
Llama-3.1-FT & \textbf{0.565} & \textbf{0.130} & \textbf{0.018} & 0.830 & \textbf{0.120} & \textbf{0.012} & 0.602 & 0.186 & 0.193 & 0.517 & \textbf{0.161} & \textbf{0.004} & 0.577 & \textbf{0.163} & 0.077 \\
 \bottomrule
\end{tabular}
}
\caption{Performance of fine-tuned Llama-3.1 model for different languages. 
Llama-3.1: Llama-3.1-8B-Instruct, Llama-3.1-FT: Fine-tuned.
}
\label{tab:ft_results}
\vspace{-0.3cm}
\end{table*}

\section{Results}
\label{sec:results}


\noindent\textbf{Open {\em vs} Close LLMs.}
We report the performance of both open- and closed-LLMs across all the regions in Table \ref{tab:results}. Our results indicate that the closed models (e.g., GPT-4o BLEU-AVG:0.230), outperform the open models (LLama3.1 BLEU-AVG:0.186) significantly. Within the closed models, Gemini performs better in terms of semantic measure, in most of the regions, with GPT4o closely following.  Llama3.1 leads the open models in both the lexical and semantic measures across majority of the regions.

\noindent\textbf{High- {\em vs} Low-resource Languages.} Figure \ref{fig:bleu_score_plot} reports the average BLEU scores across all the regions, grouped by the four resource tiers: high- to extremely-low resource languages. We find that L2 English achieves the highest performance, while Assamese has the lowest.
This clearly indicates that the performance correlates to the representation and/or richness of digital content of the language used in the models. 


\noindent\textbf{Fine-tuned Models.}
Our findings, reported in Table \ref{tab:ft_results}, indicate that fine-tuning with the \mnqa{} train set mostly improves performance for (extremely-)low resource language such as Assamese and Nepali. For the medium resources, the results are mixed. We observe that fine-tuning benefits dialect-rich languages (e.g., Arabic) more than similarly resourced ones, likely due to native datasets enhancing cultural and dialectal knowledge. For high-resource languages, the fine-tuned model largely retains the base model's strengths.


 
\noindent\textbf{LLM-as-a-judge.}
The performance of the LLM-as-a-judge approach is presented in Table \ref{tab:model_performance-llm-as-a-judge}. 
Our findings align with other evaluation metrics, showing that high-resource languages (e.g., En) perform 
better than low-resource languages (e.g., Asm).

\begin{table}[h]
    \centering
\setlength{\tabcolsep}{2pt} 
\scalebox{0.9}{%
    \begin{tabular}{lrrrrr}
        \toprule
        \textbf{Language} & \textbf{GPT-4o} & \textbf{Gemini} & \textbf{Llama} & \textbf{Mistral} & \textbf{Avg.} \\
        \midrule
        Arabic        & 6.03  & 6.39  & 4.27  & 3.79  & \textbf{5.12} \\
        Assamese      & 4.82  & 4.17  & 2.71  & 2.31  & \textbf{3.50} \\
        Bangla-BD     & 5.08  & 5.32  & 3.11  & 1.53  & \textbf{3.76} \\
        Bangla-IN     & 5.71  & 6.03  & 3.63  & 2.52  & \textbf{4.47} \\
        English-BD    & 6.33  & 6.64  & 6.30  & 5.34  & \textbf{6.15} \\
        English-QA    & 6.16  & 6.57  & 6.24  & 5.49  & \textbf{6.12} \\
        Hindi         & 6.87  & 7.22  & 5.28  & 4.87  & \textbf{6.06} \\
        Nepali        & 5.68  & 6.26  & 3.53  & 1.34  & \textbf{4.20} \\
        Turkish       & 5.51  & 4.51  & 4.05  & 2.36  & \textbf{4.11} \\
        \midrule
        \textbf{Average} & \textbf{5.80} & \textbf{5.90} & \textbf{4.35} & \textbf{3.28} &  \\
        \bottomrule
    \end{tabular}
    }
    \caption{Performance of all LLMs evaluated using GPT-4o as a judge across languages. `Gemini' refers to Gemini 1.5, `Llama' to Llama 3.1 8b, and `Mistral' to Mistral 7b. Responses were rated on a scale of 1 to 10, with higher scores indicating better performance.}
    \label{tab:model_performance-llm-as-a-judge}
    \vspace{-0.2cm}
\end{table}

\noindent\textbf{Subjective Evaluation.}
We performed qualitative evaluation of GPT$-4o$ model for 
all languages except Hindi and Nepali. 
For the qualitative analysis, we sampled 100 QA pairs from each languages and observed an average accuracy rating of 4.08 (out of 5) and average usefulness of 4.02 (/5). The results are presented in Table \ref{tab:human_evaluation_scores}. See Sec. \ref{sec:app_subjective_evaluation} for evaluation criteria.
Our error analysis highlights three key issues: \textit{(i)} inaccuracies in answers to ``proper noun'' questions requiring region-specific responses (e.g., India); \textit{(ii)} difficulty answering questions related to the current year (2024); and \textit{(iii)} errors in numerical questions requiring precise values. 
Detailed examples are in Appendix Figure \ref{fig:app_error_analysis_as_bn} and \ref{fig:app_error_analysis_hi}.

 \begin{table}[h]
    \centering
\setlength{\tabcolsep}{2pt} 
\scalebox{0.70}{%
    \begin{tabular}{l r r r r r r r r}
        \toprule
        \textbf{Metrics}  & \textbf{Ar} & \textbf{As} & \textbf{Bn(BD)} & \textbf{Bn(IN)} & \textbf{En(BD)} & \textbf{En(QA)} & \textbf{Tr} & \textbf{Avg.} \\
        \midrule
        Accuracy   & 4.56 & 3.86 & 3.41 & 3.49 & 4.57 & 4.91 & 3.82 & \textbf{4.09}\\
        Usefulness & 4.55 & 3.80 & 3.40 & 3.46 & 4.63 & 4.91 & 3.45 & \textbf{4.03}\\
        \bottomrule
    \end{tabular}
    }
    \caption{Human evaluation scores on a Likert scale (1–5) for accuracy and usefulness across all languages, except Hindi and Nepali. Assessed on a Likert scale (1–5), higher is better.}
    \label{tab:human_evaluation_scores}
    \vspace{-0.4cm}
\end{table}

\section{Conclusions}
\label{sec:conclusions}
In this paper, we propose the \nqa{} framework, to enable constructing culturally and regionally-aligned natural QA datasets with minimal human-effort. The proposed framework is scalable and language-independent, which not just facilitate creating region- and culture-based benchmarking efforts, but also resources that can be used in continual learning or fine-tuning the LLMs. We show the efficacy of the \nqa{}, by designing and developing a multilingual native QA dataset, \mnqa{} -- from $9$ regions ($7$ languages) encapsulating the scenario of high-low resource representation. We benchmark the \mnqa{} with 2 open and 2 closed LLMs. Our results indicate the superiority of closed models over open LLMs, and the performance gaps between high- and low-resource languages. By utilizing the \mnqa{} dataset for fine-tuning, we can potentially inject cultural and regional knowledge into the LLMs, as evidenced by the improved performance of Arabic, a mid-resource language, and Assamese, an extremely low-resource language. Our future work includes extending the \nqa{} framework with additional search engine capabilities, image and video search options, and releasing it to the community for seamless use in research~\cite{alam2025nativqaframework}. 

\section{Limitations}
\label{sec:limitations}
While the proposed framework enables the development of datasets with cultural and native information, it currently has several limitations. Firstly, the \nqa{} framework relies on human-in-the-loop processes, from seed query creation to manual revision of QA pairs. This dependency limits large-scale data collection. Although we consider the human-in-the-loop setting a limitation, we also note that ensuring a high-quality dataset without it would be challenging. Secondly, the semi-supervised approach, which is based on domain reliability checking (DRC) is a reasonable starting point; however, full supervision would ensure higher quality.

\section*{Ethics Statement}
\label{sec:ethics}

The proposed \nqa{} does not involve collecting any personally identifiable information. Additionally, the proposed dataset does not include any information that can offend or harm any individual, entity, organization, or society. Therefore, we do not foresee any potential risk.

\bibliography{bib/bibliography}

\section*{Appendix}
\label{sec:appendix}
\appendix

\section{Related Existing Work}
\label{sec:app_existing_work}
In Table \ref{tab:existing_data}, we present a comparison with previous work, highlighting how the \mnqa{} dataset differs from prior studies. 

\begin{table*}[h]
\centering
\setlength{\tabcolsep}{2pt} 
\scalebox{0.8}{%
\begin{tabular}{@{}lcccr@{}}
\toprule
\multicolumn{1}{c}{\textbf{Dataset}} & \multicolumn{1}{c}{\textbf{\# of Lang}} & \multicolumn{1}{c}{\textbf{Lang}} & \multicolumn{1}{c}{\textbf{Domain}} & \multicolumn{1}{c}{\textbf{Size}} \\ \midrule
SquAD \cite{rajpurkar-etal-2016-squad} & 1 & En & Wiki & 100K \\
TriviaQA \cite{joshi-etal-2017-triviaqa} & 1 & En & Wiki, Web & 650K \\
HotpotQA \cite{yang-etal-2018-hotpotqa} & 1 & En & Wiki & 113K \\
NQ \cite{kwiatkowski-etal-2019-natural} & 1 & En & Wiki & 323K \\
XQA~\cite{liu-etal-2019-xqa} & 9 & \begin{tabular}[c]{@{}l@{}} En, Zh, Fr, De, Pl, Pt, Ru, Ta, Uk \end{tabular}  & Wiki & 90K \\
TyDiQA~\cite{clark2020tydi} & 11 & En, Ar, Bn, Fi, Id, Ja, Sw, Ko, Ru, Te, Th   & Wiki  & 204k  \\
GooAQ~\cite{khashabi-etal-2021-gooaq-open} & 1  & En  & Open  & 3M \\
BanglaRQA~\cite{ekram-etal-2022-banglarqa} & 1 & Bn & Wiki & 3k \\
HelpSteer \cite{wang2023helpsteer} & 1 & En & Helpfulness & 37K \\
BLEnD~\cite{myung2024blend} & 13 & \begin{tabular}[c]{@{}l@{}}En, Zh, Es, Id, Ko, El, Fa, \\Ar, Az, Su, As, Ha, Am \end{tabular} & Open & 52.5k \\
CaLMQA~\cite{arora2024calmqa} & 23 & \begin{tabular}[c]{@{}l@{}} En, Ar, Zh, De, Hi, He, Hu, Ja, Ko, Es, Ru, Aa, \\Bal, Fo, Fj, Hil, Rn, Pap, Ps, Sm, To, Tn, Wol \end{tabular}  & Open & 1.5K \\
\rowcolor{lime} \mnqa{} dataset & 7 & \begin{tabular}[c]{@{}l@{}}Ar, As, Bn, \\En, Hi, Np, Tr\end{tabular} & Open & $\sim$64K \\ 
\bottomrule
\end{tabular}
}
\caption{The most notable existing QA datasets compared to \mnqa{}.}
\label{tab:existing_data}
\vspace{-0.4cm}
\end{table*}

\section{Query on Search Engine}
\label{sec:app_query_search_engine}
\begin{figure}[t]
    \centering    
    \frame{\includegraphics[scale=0.3]{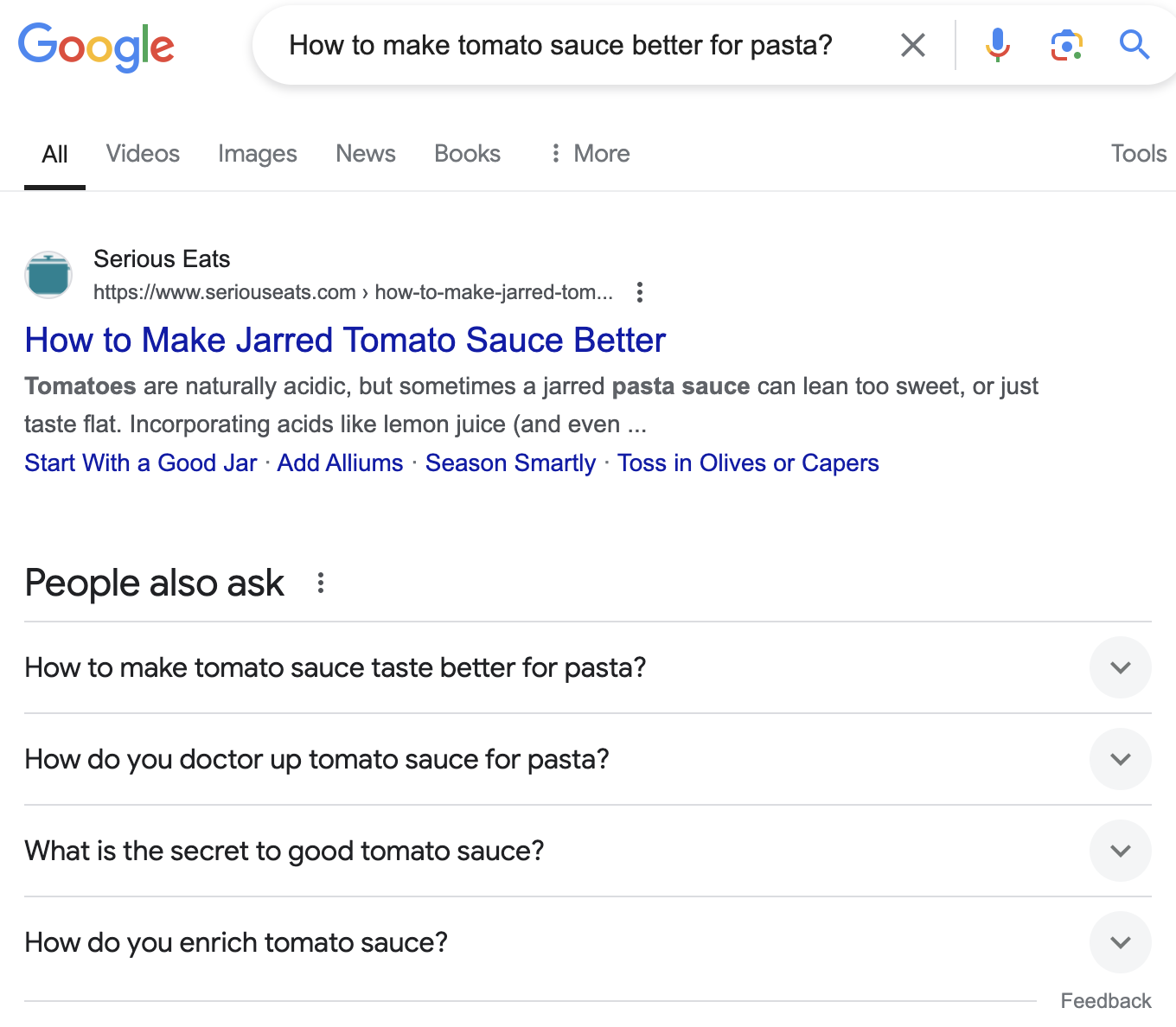}}        
    \caption{An example of search interface showing search response with \textit{``people also ask''} option.}
    \label{fig:search_interface}
\end{figure}

In Figure \ref{fig:search_interface}, we show an example of a query to a search engine that demonstrates related queries under ``People also ask'', which we have also considered as queries in the several iterations of QA pair collection.




\section{Detailed Annotation Guideline}
\label{sec:det_annot_guidelines}

\subsection{Collecting Seed Queries}
\label{sec:app_seed_queries}

The purpose of this study is to collect natural QA pairs to evaluate and enhance LLMs. Our approach to collecting such QA pairs is to utilize widely used search engines with natural queries to find relevant QA pairs. We intended to find a diverse set of questions; therefore, we selected 18 different topics as discussed in Section \ref{ssec:qc}.


For each topic, the task was to collect seed queries. While collecting the seed queries, we needed to ensure language-specific and major-city-centric information as naturally as possible, information we typically ask on search engines. For example, \textit{``Does Qatar have beaches?''} or \textit{``Do I need a visa to visit Qatar?''}

These examples are based on Qatar; however, for each language, the questions will be specific to the specified location (major city/country).

\subsection{Domain Reliability}
\label{sec:app_domain_reliablity}



For the domain reliability task annotators were tasked to review each web domain to determine its credibility and assign one of the following four reliability labels:
\noindent\begin{itemize}[noitemsep,topsep=0pt,leftmargin=*,labelsep=.5em]
    \item \textbf{Very reliable:} The information is accepted without additional verification.
    \item \textbf{Partially reliable:} The information may need further verification.
    \item \textbf{Not sure:} Unable to verify or judge the website for any reason.
    \item \textbf{Completely unreliable:} The website and the information appear unreliable.
\end{itemize}

\paragraph{General Characteristics}
Below are some characteristics that we have considered as criteria for a domain to be considered more reliable \cite{schwarz2011augmenting,flanagin2007role,metzger2015psychological,meriam2010evaluating,selejan2016credibility}.

\noindent \paragraph{Overall Design:}
\begin{itemize}[noitemsep,topsep=0pt,leftmargin=*,labelwidth=!,labelsep=.5em]
     \item The domain has a professional, polished, and attractive design. It has interactive features, is well organized, easy to navigate, loads fast, and has good response speed. 
     \item There are no errors or broken links.
     \item It might have paid access to information.
     \item The domain name suffix is considered trustworthy (e.g., ``.gov'').
     \item Absence/limited advertising. If advertisements are present, they are good quality ads for reputable and decent products and organizations.
     \item The domain might be sponsored by or shows links to reputable organizations.
     \item Presence of a section or page on privacy and security, About page, contact info, and address.  
     \item If videos, images, and graphics are used on the website, they are high-quality and professional.       
\end{itemize}

\noindent \paragraph{Content Quality:}
\begin{itemize}[noitemsep,topsep=0pt,leftmargin=*,labelwidth=!,labelsep=.5em]
    \item Author/entity names, qualifications, credentials, and contact information are present, and they are relevant to the topic of the website or the content presented. 
    \item Author/entity is reputable. 
    \item Contains date stamp.
    \item Presents information that is current and up to date. 
    \item Has citations, especially to scientific data or references, and shows links to external authorities.
    \item Content is relevant to the target topic and current events.
    \item Professional-quality, clear writing, and good formatting of text.
    \item Content appears accurate, lacks bias, factually correct, plausibility, and uses appropriate objective language. 
    \item Free of misspellings and grammar mistakes.
    \item The information provided is at an appropriate level, not too generic or elementary. 
\end{itemize}     

\noindent \paragraph{General Instructions:} We also provided the following general instructions to guide annotators. 
\begin{itemize}[noitemsep,topsep=0pt,leftmargin=*,labelwidth=!,labelsep=.5em]
    \item Do not spend more than five minutes per given Web domain. 
    \item Explore/observe/look at \textbf{ALL} elements in the domain’s home page from top to bottom.
    \item Repeat points 1-2 on other pages from the same domain, and look at their content, structure, design, author, etc. \textit{You are not required to read these pages in full, reading the first 1-2 paragraphs is enough.}
    \item During annotation, consider the annotation criteria mentioned in this guideline, and evaluate each source based on those aspects. A ``reliable website'' might not meet all those criteria. It is your job, as annotator, to measure the website's reliability guided by these criteria.
    \item You should evaluate a domain based on what is presented on it only. You should not navigate or search in outside sources, even if some are linked inside the given domain/page.
    \item Please use ``Not sure'' very sparingly in rare cases when you are extremely unsure. It is preferable to always choose one of the other three labels.
    \item   For social media websites (e.g., X, Facebook) choose: Very Reliable. 
    \item For shopping websites, use the criteria listed in this guideline to decide. Some shopping websites are very reliable. 
    \item For famous people's websites, use the criteria listed in this guideline to decide.
    \item Websites that are in any other language ONLY (for example, only in English when you are working on Bangla queries), for such cases choose: Not Sure. 
\end{itemize}


\subsection{QA Annotation (Detailed Annotation Guideline)}
\label{ssec:app_answer_categorization}

\subsubsection{Question Validation:}

In this task, a pair of a question and a possible answer for that question is shown. Relying only on the question shown on the interface, the annotator is asked to perform the following tasks:

\begin{enumerate}
    \item Categorize the question as “Good” or “Bad”. Steps 2- 4 will be performed only for questions labelled as ``good''.
    \item Identify if the question is relevant to the specified location.
    \item Categorize the answer.
    \item Edit the answer (if needed). 
\end{enumerate}

The annotators classified whether the questions are ``Good'' or ``Bad'' based on the criteria discussed below. The choice of the two types of questions was inspired by the NQ dataset~\citep{kwiatkowski-etal-2019-natural}.

\begin{itemize}[noitemsep,topsep=0pt,leftmargin=*,labelsep=.5em]
    \item \textbf{Good question:} 
Is a fact-seeking question that can be answered with a name of an entity (person, place, thing.etc.), or an explanation, or a number. For examples, see Table \ref{tab:examples_1}. 
    \item \textbf{Bad question:} A question that meets any of the following criteria mentioned below. 
\end{itemize}
    
\begin{table}[htbp]
\centering
\resizebox{.45\textwidth}{!}{%
\begin{tabular}{p{0.8cm}|p{9.4cm}c}
\hline
\textbf{Lang.} & \textbf{Example} \\ \hline
\textbf{En} & Is Al Wakrah Beach free? \\
            & Do you have to pay for school in Qatar? \\ \hline
\textbf{Ar} & \begin{RLtext}\footnotesize  كم اسعار الشقق في الدوحة؟\end{RLtext} 

(\textbf{Translation:} How much is apartment rent in Doha?)
\\
            & \begin{RLtext}\footnotesize  كيف احصل على فرصة عمل في قطر؟\end{RLtext}
            (\textbf{Translation:} How do I find a job opportunity in Qatar?  )  
            \\
          
            & \begin{RLtext}\footnotesize  كيف اقدم على وظيفة في وزارة الداخلية؟\end{RLtext} 

            (\textbf{Translation:} How do I apply for a job in Ministry of Foreign Affairs?)  
            \\ \hline
\end{tabular}%
}
\caption{Examples of good questions in English and Arabic.}
\label{tab:examples_1}
\end{table}

\begin{itemize}
    \item It is unclear what the question is asking for. See Table \ref{tab:examples_2}. 
     \item Incomprehensible as a result of grammatical errors. This will be a rare case. Some grammatical mistakes can be acceptable as long as its meaning is understandable.  
     \item The question depends on clear false presupposition, such as racist, sexist, or stereotypical ideas as shown in Table \ref{tab:controversial_questions}. 
    \item The question is opinion-seeking. It asks the public opinion on a topic/matter where finding a concrete answer is difficult. For examples, see Table \ref{tab:questions_examples_3}. 
    \item The question does not clearly ask for factual information. See Table \ref{tab:questions_examples_4}. 
\end{itemize}

\begin{table}[htbp]
\centering
\renewcommand{\arraystretch}{1.2}
\scalebox{0.8}{%
\begin{tabular}{p{2cm}|p{7cm}}
\hline
\textbf{Language} & \textbf{Example} \\ \hline
\textbf{En} & Is Indonesian coffee strong? \\ \hline
\textbf{As/Bn/Hi} (Indic Languages) & 
\textbf{Note: Special cases only for India (Indic Languages)} \newline
- Who is the chief minister of India? \newline
(Reason: In India, the chief minister is referred to state-wise, for example, ``Who is the chief minister of West Bengal?'') \newline
- What is the national language of India? \newline
(Reason: In India, the correct question would be, “What is the official language of India?” As per the constitution, Hindi and English are the official languages, and 22 languages are recognized as per the Eighth Schedule of the Constitution. India has no national language.) \newline
- What is the national animal of Assam/West Bengal? \newline
(Reason: National animals refer to countries, not states. For example, ``What is the national animal of India?'') \\ \hline
\textbf{Ar} & \begin{RLtext}\footnotesize  كيف اكتب كاس العالم؟\end{RLtext}

(\textbf{Translation:} How do I write ``World Cup''?)  
\\ \hline
\end{tabular}
}
\caption{Examples of bad questions in English, Indic languages, and Arabic.}
\label{tab:examples_2}
\end{table}

\begin{table}[htbp]
\centering
\renewcommand{\arraystretch}{1.2}
\scalebox{0.8}{%
\begin{tabular}{p{1cm}|p{7.5cm}}
\hline
\textbf{Lang.} & \textbf{Example} \\ \hline
\textbf{En} & 
- What is the reason that African Americans are mostly criminals? \newline
- Why does Islam encourage terrorism? \newline
- Why shouldn't women be allowed to work? \\ \hline
\textbf{Ar} & 
 \begin{RLtext}\footnotesize  
- لما معظم الأفريقيين الأمريكيين مجرمين؟ 
\end{RLtext}

(\textbf{Translation:} Why do most African American are criminals?) 
\\ &
\begin{RLtext}\footnotesize
- لماذا يشجع الإسلام على الإرهاب؟  
\end{RLtext}
(\textbf{Translation:} Why does Islam encourage terrorism? )  
\begin{RLtext}\footnotesize

\\
- لماذا يجب منع النساء من العمل؟
 \end{RLtext}
 
 (\textbf{Translation:} Why women should not allowed to work? )  

 \\ \hline
\end{tabular}
}
\caption{Examples of bad questions that depends on false presuppositions in English and Arabic.}
\label{tab:controversial_questions}
\end{table}

\begin{table}[htbp]
\centering
\renewcommand{\arraystretch}{1.2}
\scalebox{0.8}{%
\begin{tabular}{p{1cm}|p{7.5cm}}
\hline
\textbf{Lang.} & \textbf{Example} \\ \hline
\textbf{En} & 
- Can you give me your thoughts on smoking?  \newline
- Is marriage good or bad? \\ \hline
\textbf{Ar} & 
\begin{RLtext}\footnotesize  
- هل من الضروري ارتداء الزي المدرسي؟ \\
\end{RLtext} 

(\textbf{Translation:} Is it important to wear a school uniform?)  
\\ \hline
\end{tabular}
}
\caption{Examples of bad questions in English and Arabic.}
\label{tab:questions_examples_3}
\end{table}

\begin{table}[htbp]
\centering
\renewcommand{\arraystretch}{1.2}
\scalebox{0.8}{%
\begin{tabular}{p{1cm}|p{7.5cm}}
\hline
\textbf{Lang.} & \textbf{Example} \\ \hline
\textbf{En} & 
- How do you ensure you are culturally competent? \newline
- Why is it a must to preserve our local literature? \\ \hline
\textbf{Ar} & 
\begin{RLtext}\footnotesize  
- هل من السهل ايجاد عمل في قطر؟

\end{RLtext}

(\textbf{Translation:} Is it easy to find job in Qatar? )  

\\
&
\begin{RLtext}\footnotesize

- كم يستغرق الطلب تحت الاجراء قطر؟
\end{RLtext} 

(\textbf{Translation:} How long does "in process" take Qatar? )  
\\ \hline
\end{tabular}
}
\caption{Examples of bad questions in English and Arabic.}
\label{tab:questions_examples_4}
\end{table}

\subsection{Question's relevancy to the location} 

For questions labelled as ``Good'', the annotator is asked to identify whether the question is related to the specified [LOCATION]. Please see the examples below. For this step, one of the below labels should be chosen: 
\begin{itemize}
\item\textbf{Yes:} The question specifically relates to the location. For examples, see Table \ref{tab:questions_examples_relavence_1}.

\item \textbf{No:} The question is not related to the specified location, but could be related to a different location. See Table \ref{tab:questions_location}.

\item \textbf{Maybe:} The question is somewhat generic. It could apply to the specified location, but it might also be relevant to other locations. For examples, see Table \ref{tab:questions_relavence_maybe}.

\item \textbf{Unsure:} It is challenging to determine if the question is location-specific. This option should be chosen only for particularly difficult cases. For examples, see Table \ref{tab:questions_relavence_unsure}. 
\end{itemize}

\begin{table}[h]
\centering
\renewcommand{\arraystretch}{1.2}
\scalebox{0.8}{%
\begin{tabular}{p{1cm}|p{7.5cm}}
\hline
\textbf{Lang.} & \textbf{Example} \\ \hline
\textbf{En} & What is the main city in Qatar? \\ \hline
\textbf{Ar} & 
\begin{RLtext}\footnotesize  
هل قطر لديها ملك؟ 
\end{RLtext}
\rebuttal{\textbf{Translation:} Does Qatar have a king?}  

\\
&
\begin{RLtext}\footnotesize
كم عدد المساجد في دولة قطر؟
\end{RLtext} 
\rebuttal{\textbf{Translation:} How many mosques are there in Qatar?} 

\\ \hline
\end{tabular}
}
\caption{Examples of questions in English and Arabic.}
\label{tab:questions_examples_relavence_1}
\end{table}

\begin{table}[h]
\centering
\renewcommand{\arraystretch}{1.2}
\scalebox{0.8}{%
\begin{tabular}{p{1cm}|p{7.5cm}}
\hline
\textbf{Lang.} & \textbf{Example} \\ \hline
\textbf{En} & Why do Emirati men wear white robes? (the specific location was Qatar) \\ \hline
\textbf{Ar} & 
\begin{RLtext}\footnotesize 
ما هي اقامة مستثمر في السعودية؟ \end{RLtext} 
\rebuttal{\textbf{Translation:} What is investor residency is Saudi Arabia?}  
\begin{RLtext}\footnotesize

\\
(الموقع المطلوب كان قطر)
\end{RLtext} 
\rebuttal{\textbf{Translation:} The specified location in Qatar.}  
\\ \hline
\end{tabular}
}
\caption{Examples of questions in English and Arabic with specific locations.}
\label{tab:questions_location}

\end{table}

\begin{table}[h]
\centering
\renewcommand{\arraystretch}{1.2}
\scalebox{0.8}{%
\begin{tabular}{p{1cm}|p{7.5cm}}
\hline
\textbf{Lang.} & \textbf{Example} \\ \hline
\textbf{En} & 
- What is the most visited mall? \newline
- What is a place where bread and cakes are sold? \\ \hline
\textbf{Ar} & 
\begin{RLtext}\footnotesize 
- كم عدد كليات الطب؟ 
\end{RLtext}
\rebuttal{\textbf{Translation:} How many medical colleges?}  
\\&
\begin{RLtext}\footnotesize
- كم الدرجة المطلوبة في اختبار الايلتس؟
\end{RLtext}
\rebuttal{\textbf{Translation:} What is the required grade for ILETS?}  

\\ \hline
\end{tabular}
}
\caption{Examples of generic questions in English and Arabic.}
\label{tab:questions_relavence_maybe}

\end{table}

\begin{table}[h]
\centering
\renewcommand{\arraystretch}{1.2}
\scalebox{0.8}{%
\begin{tabular}{p{1cm}|p{7.5cm}}
\hline
\textbf{Lang.} & \textbf{Example} \\ \hline
\textbf{En} & 
- Is DoorDash cheaper or Uber Eats? \newline
- What are common names for Paspalum? \\ \hline
\textbf{Ar} & 
\begin{RLtext}
- كيف تعرف الصقر وهو في الجو؟ 
\end{RLtext}
\rebuttal{\textbf{Translation:} How to know the falcon while he is in the air?}  
\\& \begin{RLtext}

- ما معنى اسم عطشان؟
\end{RLtext}
\rebuttal{\textbf{Translation:} What is the meaning of the name ``Thirsty''?} 

\\ \hline
\end{tabular}
}
\caption{Examples of questions in English and Arabic.}
\label{tab:questions_relavence_unsure}

\end{table}

\subsection{Answer categorization: }
The answer of the given question should be classified using one of the below categories. The source Web page provided on the interface should be used to make the judgment. 
\begin{itemize}[noitemsep,topsep=0pt,leftmargin=*,labelsep=.5em]
    \item \textbf{Correct answer:} When the answer aligns with the information provided by the source. Note that the answer must be complete and addresses all parts of the question, but it does not need to match the source webpage verbatim. The answer can be a long, detailed response, or a short snippet. 
    \item \textbf{Partially correct answer:} When the answer does not address all parts of the question. In this case, the answer should be edited using information from the source page. The required information can be directly copied from the source webpage. Minimal editing may be needed to make the answer more comprehensive. For example, see Table \ref{tab:qa_examples_editing1}.
    \item \textbf{Incorrect answer:} When the answer does not address the question at all. In this case, the answer should be edited using information from the source page. See Table~\ref{tab:qa_examples_editing2}.
    \item \textbf{Cannot find answer:} When the answer is not available in the provided link/page, and thus, cannot be judged. 
\end{itemize}

\begin{table*}[h]
\centering

\renewcommand{\arraystretch}{1.2}
\scalebox{0.8}{%
\begin{tabular}{p{1cm}|p{5cm}|p{10cm}}
\hline
\textbf{Lang.} & \textbf{Question} & \textbf{Answer} \\ \hline
\textbf{En} & 
How many Americans live in Qatar? & 
In recent years, this figure has more than doubled and various estimates now put the number of Americans in Qatar to be up to 15,000. Most Americans within the country tend to be based in the capital city of Doha and are largely attracted by the tax-free inducement of the Persian Gulf state. \\ \hline
\textbf{AR} & 
\begin{RLtext}
من أكبر البحرين أو قطر؟
\end{RLtext} 
(\textbf{Translation:} Which is bigger: Bahrain or Qatar? )  
& 
\begin{RLtext}\footnotesize
تتنوع مساحة الدول العربية بشكل كبير، حيث تبلغ مساحة أكبر دولة عربية، وهي الجزائر، 2,381,741 كيلومتر مربع، بينما تبلغ مساحة أصغر دولة عربية، وهي البحرين، 785 كيلومتر مربع، وفقا لآخر تحديث لموقع 
\end{RLtext} worldometers.

\rebuttal{\textbf{Translation:} The area of the Arab countries varies greatly, as the area of the largest Arab country, Algeria, is 2,381,741 square kilometers, while the area of the smallest Arab country, Bahrain, is 785 square kilometers, according to the latest update to the website Worldometers.}  
\begin{RLtext}\footnotesize
\end{RLtext} \\ \hline
\end{tabular}
}
\caption{Examples of questions and answers in English and Arabic. The answers provide more information and should be edited.}
\label{tab:qa_examples_editing1}
\end{table*}

\textbf{Answer editing: } For the cases that require the answers to be edited, the below instructions should be followed: 
\begin{itemize}[noitemsep,topsep=0pt,leftmargin=*,labelsep=.5em]
    \item The parts that completely answer the question should be copied from the webpage and pasted in the answer box on the interface. This could be a long paragraph or a short snippet, or runs through multiple paragraphs.
    \item Sometimes answers may end with: (…), in such cases, the answer should be completed by finding the remaining part of the answers in the webpage.
    \item The answer should be to the point and concise. For example, if the question asks for the colour of a flag, then the answer should only answer that. Any unnecessary parts should be removed. 
\end{itemize}

\begin{table*}[h]
\centering
\renewcommand{\arraystretch}{1.2}
\scalebox{0.8}{%
\begin{tabular}{p{1cm}|p{6cm}|p{8cm}}
\hline
\textbf{Lang.} & \textbf{Question} & \textbf{Answer} \\ \hline
\textbf{En} & 
Does Qatar have online shopping? & 
Carrefour Qatar - Shop Online for Grocery, Food, Mobiles, Electronics, Beauty, Baby Care \& More. \\ \hline
\textbf{Ar} & 
\begin{RLtext}
من هي اغنى عائلة في قطر؟
\end{RLtext}
\rebuttal{\textbf{Translation:} Who is the richest family in Qatar?} 

& 
\begin{RLtext}
جاءت عائلة ساويرس في المرتبة الأولى كأغنى عائلة في المنطقة العربية، بصافي ثروة إجمالية قدرها 11.2 مليار دولار.
\end{RLtext}
\rebuttal{\textbf{Translation:} The Sawiris family ranked first as the richest family in the Arab region, with a total net worth of 11.2 billion dollar.}  

\\ \hline
\end{tabular}
}
\caption{Examples of questions and wrong answers in English and Arabic. The answers need to be edited.}
\label{tab:qa_examples_editing2}
\end{table*}

\subsection{Annotation Platform} 
\label{sec:app_annotation_platform}
We utilized in-house annotation platform for the tasks. Separate annotation interfaces (as presented in Appendix \ref{sec:app_annotation_interface}) were designed for each phase and each language, resulting 18 annotation projects. To facilitate the annotation process, the annotation interface included the annotation guidelines throughout the phases.

\section{Additional Statistics}
We computed the average length of questions and answers for each language, where word boundaries were identified using whitespace tokenization. We use white spaces as the word boundaries. A breakdown of the average lengths per language is provided in Table \ref{tab:qa_avg_length}.

\begin{table}[tbh!]
    \centering
\setlength{\tabcolsep}{2pt} 
\scalebox{0.9}{%
    \begin{tabular}{l rr}
        \toprule
        \textbf{Lang} & \textbf{Question (Avg)} & \textbf{Answer (Avg)} \\
        \midrule
        Arabic      & 6.0  & 35.1 \\
        Assamese    & 6.0  & 34.6 \\
        Bangla-BD   & 6.1  & 34.9 \\
        Bangla-IN   & 5.4  & 31.9 \\
        English-BD  & 6.2  & 34.6 \\
        English-QA  & 6.4  & 36.4 \\
        Hindi       & 6.4  & 36.3 \\
        Nepali      & 6.4  & 36.3 \\
        Turkish     & 6.2  & 35.4 \\
        \bottomrule
    \end{tabular}
    }
    \caption{Average length (in words) of questions and answers per language.}
    \label{tab:qa_avg_length}
\end{table}

\section{Prompting and Instruction Tuning: Additional Details}
\label{sec:app:prompting}
\subsection{Prompts}
\label{sec:app:prompting_zshot}
In our main experiments of zero-shot prompting of the different LLMs, we manually and carefully designed a prompt to instruct a model to perform the QA task. Our prompt engineering process is inspired by relevant research and our experimental observations over the development sets. For this experiment, we use the system and user prompts reported in Table~\ref{tab:prompts_main}.

\begin{table*}[h]
\centering
\setlength{\tabcolsep}{2pt} 
\scalebox{0.9}{%
\begin{tabular}{lp{13cm}}
\toprule
\textbf{Role} & \textbf{Prompt}  \\
\midrule
System & You are a/an \textbf{[\textit{lang}]} AI assistant specializing in both short and long-form question answering. Your task is to provide clear, accurate, and relevant responses across various fields, ensuring concise and well-structured answers.  \\
\midrule
User  & Please use your expertise to answer the following \textbf{[\textit{lang}]} question. Answer in \textbf{[\textit{lang}]} and rate your confidence level from 1 to 10. Provide your response in the following JSON format: \{``answer'': ``your answer'', ``score'': your confidence score\}. Please provide JSON output only. No additional text. \textbf{Question}: input\_question\\
\bottomrule
\end{tabular}
}
\caption{Prompts used with the LLMs for zero-shot question answering.  \textbf{\textit{lang}}: the language of QA pair.}
\label{tab:prompts_main}
\end{table*}

\subsection{Prompt for Query Expansion}
\label{sec:app:prompting_query_expansion}

The idea of query expansion was to create a diverse set of queries to collect more QA pairs. Table \ref{tab:prompts_sim_query_gen} presents the prompts used for query expansion with GPT-4o.

\begin{table}[h]
\centering
\setlength{\tabcolsep}{2pt} 
\scalebox{0.9}{%
\begin{tabular}{lp{7cm}}
\toprule
\textbf{Role} & \textbf{Prompt}  \\
\midrule
System & You are an expert for query expansion.  \\
\midrule
User  & For the following query, please try to expand it. Please provide output in a list in a JSON format. \\
 & Query: $input\_query$ \\
 & Expanded Queries: \\
\bottomrule
\end{tabular}
}
\caption{Prompts used to generate similar queries through GPT-4o.}
\label{tab:prompts_sim_query_gen}
\end{table}

\subsection{Instruction Generation}
\label{sec:app:prompting_inst}

To generate instruction templates through GPT-4o and Claude-3.5 Sonnet, we use the prompt in Table~\ref{tab:prompts_inst_gen}. Table~\ref{tab:ex_insts} shows examples of the generated instructions. Note that we only generate instructions for the user role, while we keep the system role fixed to that presented in  Table~\ref{tab:ex_insts}. For all generated instructions, we append the following suffix to the instruction to further instruct the LLM to comply to our requirement of concise answers: \textit{Make your answer very concise and to the point. Return only the answer without any explanation, justification or additional text}.

\begin{table*}[h]
\centering
\setlength{\tabcolsep}{2pt} 
\scalebox{0.9}{%
\begin{tabular}{lp{13.5cm}}
\toprule
\textbf{Role} & \textbf{Prompt}  \\
\midrule
System & You are an expert LLM developer with expertise in writing instructions to instruction-tune LLMs for users' tasks.  \\
\midrule
User  & We are creating an English instruction-following dataset for question answering task. An example instruction is: Interpret the following question about the real world carefully and research each answer, then provide a clear and concise answer to the question. Write 10 very diverse and concise English instructions. Only return the instructions without additional text. Return the instructions as strings in a list format as follows: {[}{]}\\
\bottomrule
\end{tabular}
}
\caption{Prompts used to generate instructions through LLMs. \label{tab:prompts_inst_gen}}

\end{table*}

\begin{table*}[t]
\centering
\setlength{\tabcolsep}{4pt} 
\scalebox{0.9}{%
\begin{tabular}{lp{5cm}p{7.5cm}}
\toprule
\textbf{Model} & \textbf{Instruction}  & \textbf{System Role}  \\
\midrule
GPT-4o & Analyze the given question thoroughly and provide a well-researched and precise answer. & You are a/an \textbf{[\textit{lang}]} AI assistant specialized in providing detailed and accurate answers across various fields. Your task is to deliver clear, concise, and relevant information.\\
\midrule
Claude-1.5  &  Carefully consider the question and provide a short, well-researched answer that covers all key points.  & You are a/an \textbf{[\textit{lang}]} AI assistant specialized in providing detailed and accurate answers across various fields. Your task is to deliver clear, concise, and relevant information.\\
\bottomrule
\end{tabular}
}
\caption{Examples of instructions generated by two LLMs along with the pre-defined system role prompt. \textbf{\textit{lang}}: the language of QA pairs for which the final instruction will be created.\label{tab:ex_insts}}
\end{table*}

\section{Dataset: Additional Data}
\label{sec:app_additional_data}
In addition to the dataset summarized in Table~\ref{tab:stats_queries}, we have collected un-annotated QA pairs for additional locations. Table~\ref{tab:additional_data} shows statistics of  collected Arabic and English data in different locations.

\begin{table}[h]
\centering
\setlength{\tabcolsep}{4pt} 
\scalebox{0.9}{%
\begin{tabular}{@{}lrlr@{}}
\toprule
\multicolumn{1}{c}{\textbf{Lang-Loc}} & \multicolumn{1}{c}{\textbf{\# of QA}} & \multicolumn{1}{c}{\textbf{Lang-Loc}} & \multicolumn{1}{c}{\textbf{\# of QA}} \\ \midrule
Ar-Egypt & 7,956 & Ar-Tunisia & 14,789 \\
Ar-Palestine & 5,679 & Ar-Yemen & 4,818 \\
Ar-Sudan & 4,718 & En-New York & 6,454 \\ \midrule
\textbf{Total} & \multicolumn{3}{r}{\textbf{55,702}} \\ \bottomrule
\end{tabular}
}
\caption{Statistics of additional QA pairs collected for different locations through our framework.}
\label{tab:additional_data}
\end{table}

\section{Annotated Dataset: Additional Details}
\label{sec:app_additional_details}

\begin{figure*}[h]
    \centering
    \begin{subfigure}{0.45\textwidth}
        \centering
        \includegraphics[width=\linewidth]{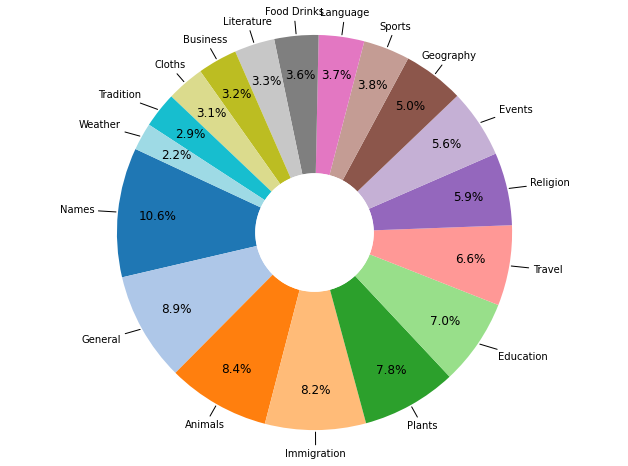}
        \caption{Arabic}
    \end{subfigure}    
    \hfill
    \begin{subfigure}{0.45\textwidth}
        \centering
        \includegraphics[width=\linewidth]{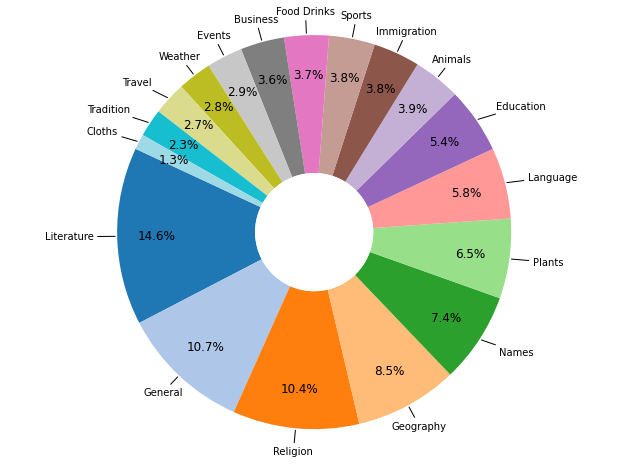}
        \caption{Assamese}
    \end{subfigure}    
    \begin{subfigure}{0.45\textwidth}
        \centering
        \includegraphics[width=\linewidth]{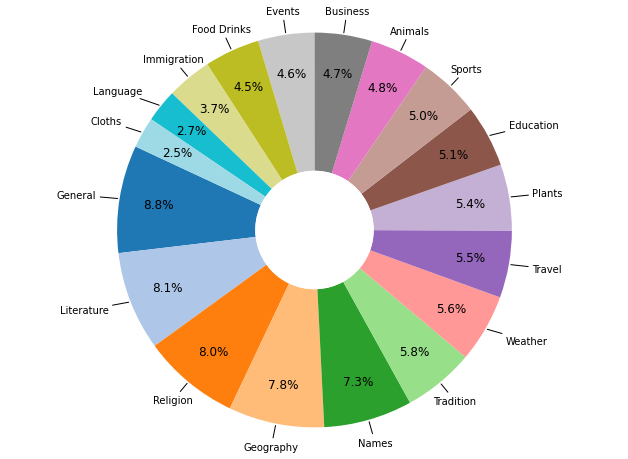}
        \caption{Bangladeshi Bangla}
    \end{subfigure}
    \hfill
    \begin{subfigure}{0.45\textwidth}
        \centering
        \includegraphics[width=\linewidth]{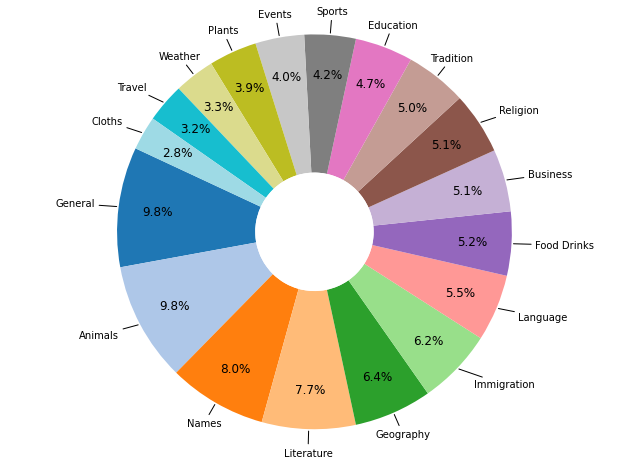}
        \caption{Indian Bangla}
    \end{subfigure}
    

    \caption{Topic wise distribution in different languages such as \textit{Arabic}, \textit{Assamese}, \textit{Bangladeshi Bangla}, and \textit{Indian Bangla}, 
    }
    \label{fig:topic_wise_dist1}
\end{figure*}

\begin{figure*}[h]
    \centering
    
    \begin{subfigure}{0.45\textwidth}
        \centering
        \includegraphics[width=\linewidth]{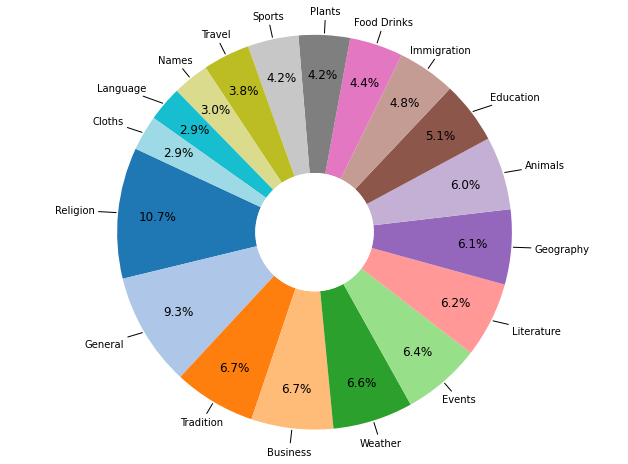}
        \caption{English in Bangladesh}
    \end{subfigure}
    \hfill
    \begin{subfigure}{0.45\textwidth}
        \centering
        \includegraphics[width=\linewidth]{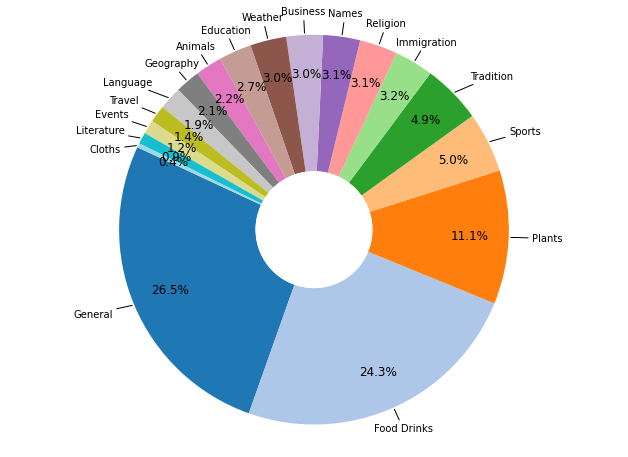}
        \caption{English in Qatar}
    \end{subfigure}    

    \caption{Topic wise distribution in different languages such as 
    \textit{English in Bangladesh}, and \textit{English in Qatar}.}
    \label{fig:topic_wise_dist2}
\end{figure*}

\begin{figure*}[t]
    \centering
    \begin{subfigure}{0.45\textwidth}
        \centering
        \includegraphics[width=\linewidth]{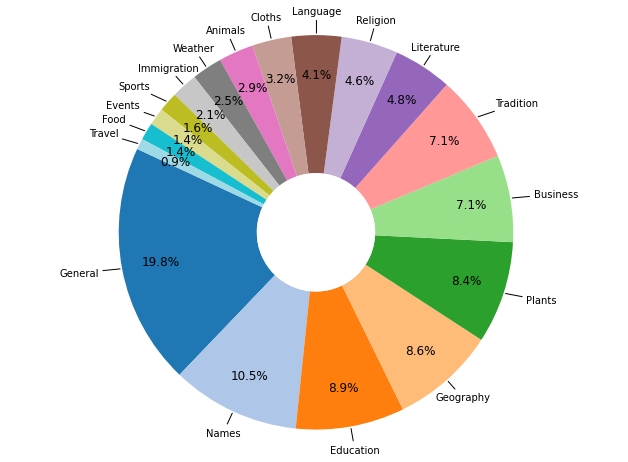}
        \caption{Nepali}
    \end{subfigure}
    \hfill
    \begin{subfigure}{0.45\textwidth}
        \centering
        \includegraphics[width=\linewidth]{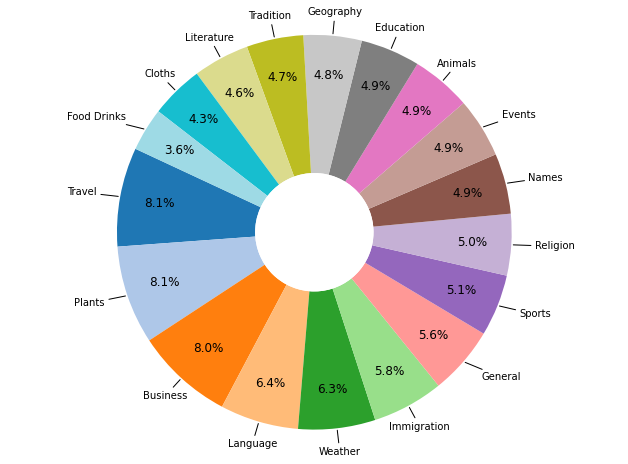}
        \caption{Hindi}
    \end{subfigure}
    \caption{Topic wise distribution for \textit{Nepali}, and \textit{Hindi}}
    \label{fig:topic_wise_dist3}
\end{figure*}

\begin{figure}[t]
    \centering
    \begin{subfigure}{0.45\textwidth}
        \centering
        \includegraphics[width=\linewidth]{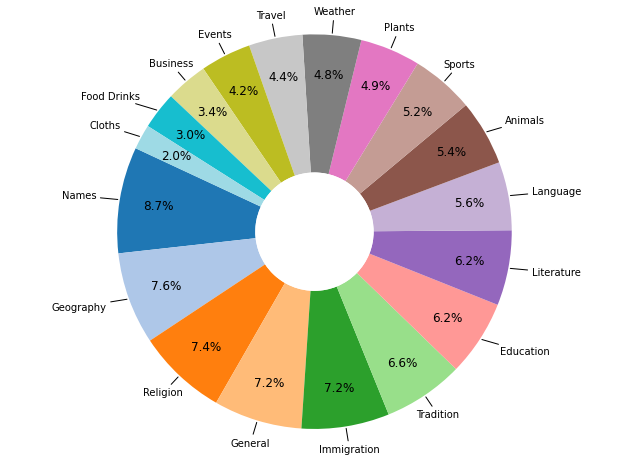}
        \caption{Turkish}
    \end{subfigure}    
    \caption{Topic wise distribution for \textit{Turkish}.}
    \label{fig:topic_wise_dist4}
\end{figure}

In Figure \ref{fig:topic_wise_dist1}, \ref{fig:topic_wise_dist2}, \ref{fig:topic_wise_dist3} and \ref{fig:topic_wise_dist4} we present the topic-wise data distribution for different datasets associated with various languages. Starting with the Arabic dataset, the predominant topic is \textit{names}, comprising 10.6\% of the data. For Assamese, the major category is Literature (14.6\%). For Bangla, whether from Bangladesh or India, the major topic is \textit{general}, representing 8.8\% and 9.8\% respectively. In Bangladesh, \textit{religion} (10.7\%) is the major topic for English, whereas in Qatar, \textit{general} dominates at 26.5\% and \textit{food and drinks} dominates a second major topic. For Nepali, the leading topic is \textit{general} (19.8\%), for Hindi it is \textit{travel} and \textit{plant} (8.1\% for each topic), and for Turkish, \textit{names} is the primary topic at 8.7\%.


\section{Dataset: Annotation (Answer Editing) Analysis}
\label{sec:data_annotation_analysis}
We computed the normalized Levenshtein distance between the original answer collected using \textit{NativQA} framework and the annotated answer to identify the robustness of \textit{NativQA} framework. During the distance computation, we provide a weight of 1 for insertion, deletion, and substitution operations. The average edits across all languages are relatively low ($0.17$), which indicates minimal edits has been done on the answers. In Table \ref{tab:app_lev_distance}, we provide distance measures for all languages across different data splits. As shown in the table, the majority of edits were made for Hindi, Nepali, and Bangla (IN), with distance measures of 0.336, 0.302, and 0.266, respectively. Overall, the edits are relatively low across languages, suggesting that the semi-supervised approach used in the \textit{NativQA} framework can be adapted for creating resources for other languages and locations. 

\begin{table*}[]
\centering
\setlength{\tabcolsep}{2pt} 
\scalebox{0.8}{%
\begin{tabular}{lrrrrr}
\toprule
\multicolumn{1}{c}{\textbf{Data Split}} & \multicolumn{1}{c}{\textbf{Arabic}} & \multicolumn{1}{c}{\textbf{Assamese}} & \multicolumn{1}{c}{\textbf{Bangla (BD)}} & \multicolumn{1}{c}{\textbf{Bangla (IN)}} & \multicolumn{1}{c}{\textbf{English (BD)}} \\ \midrule
Train & 0.196 & 0.136 & 0.191 & 0.265 & 0.114 \\
Dev & 0.063 & 0.096 & 0.307 & 0.366 & 0.160 \\
Test & 0.229 & 0.165 & 0.005 & 0.166 & 0.001 \\ \midrule
\textbf{Average} & 0.163 & 0.132 & 0.168 & 0.266 & 0.092 \\ \midrule
\multicolumn{1}{c}{\textbf{}} & \multicolumn{1}{c}{\textbf{English (QA)}} & \multicolumn{1}{c}{\textbf{Hindi}} & \multicolumn{1}{c}{\textbf{Nepali}} & \multicolumn{1}{c}{\textbf{Turkish}} & \multicolumn{1}{c}{\textbf{Average (Split)}} \\ \midrule
Train & 0.149 & 0.362 & -- & 0.052 & 0.188 \\
Dev & 0.053 & 0.186 & -- & 0.190 & 0.143 \\
Test & 0.043 & 0.460 & 0.302 & 0.186 & 0.248 \\ \midrule
\textbf{Average} & 0.082 & 0.336 & 0.302 & 0.143 & \\ 
\bottomrule
\end{tabular}
}
\caption{Normalized Levenshtein distance for all languages across different splits. \textit{Average (Split)} indicates on average distance measure across splits. $-$ No training and dev sets for Nepali.}
\label{tab:app_lev_distance}
\end{table*}

\section{Language Specific Models for BERTScore}
\label{sec:app_BERTScore_models}
In Table \ref{tab:language_models_bert_score}, we present the pre-trained language models used with BERTScore to account for language-specific variations in the evaluation measures.

\begin{table}[h]
\centering
\setlength{\tabcolsep}{2pt} 
\scalebox{0.9}{%
\begin{tabular}{l|l}
\toprule
\textbf{Lang./Region} & \textbf{Model} \\ \midrule
Arabic      & aubmindlab/bert-base-arabertv2 \\
Assamese    & ai4bharat/indic-bert \\
Bangla (BD)      & csebuetnlp/banglabert \\
Bangla (IN)      & sagorsarker/bangla-bert-base \\
English (BD)     & bert-base-uncased \\ 
English (QA)     & bert-base-uncased \\ 
Hindi    & ai4bharat/indic-bert \\ 
Nepali & bert-base-multilingual-uncased \\
Turkish & dbmdz/bert-base-turkish-cased \\ \bottomrule
\end{tabular}
}
\caption{Language specific models used to compute BERTSCore. Model id is same on HuggingFace.}
\label{tab:language_models_bert_score}
\end{table}

\section{Evaluation: LLM-as-a-judge}
\label{sec:app_eval_llm-as-a-judge}
We have computed the performance of the all models using GPT-4o-as-a-judge, following the pointwise LLM-as-judge approach with reference answers \cite{zheng2023judging}. Please find the instruction below: 

\lstset{frame=none}
\begin{lstlisting}[]
Instruction:
``Please act as an impartial judge and evaluate the quality of the response provided by AI assistant to the user question displayed below. You will be given a reference answer. Your evaluation should consider factors such as the helpfulness, relevance, accuracy, depth, creativity, and level of detail of the response. Begin your evaluation by comparing the assistant's answer with the reference answer. Then provide a short explanation. Be as objective as possible. After providing your explanation, please rate the response on a scale of 1 to 10. '''
\end{lstlisting}
Based on these results, our observation holds with other metrics - performance of high-resourced languages (e.g., English) is relatively better than low-resourced languages (e.g., Assamese). Results are reported in Table \ref{tab:model_performance-llm-as-a-judge}.  


\section{Human (Subjective) Evaluation}
\label{sec:app_subjective_evaluation}
The goal of the human evaluation task was to rate the \textit{accuracy} and \textit{usefulness} of an LLM's output. The rating scale ranges from 1 to 5, where higher values indicate better performance in both categories. We defined the measures and their guidelines as follows:

\textbf{Accuracy:} Measures whether the answer is factually correct and aligns with established knowledge or the provided context. Consider whether the answer presented is free from errors, consistent with known information, and precise in its claims. 
The rating score representing accuracy is as follows:

\begin{itemize}[label={}]
\item \textit{5: Very Accurate:} The answer is completely accurate, without any errors. All claims and facts presented are correct and aligned with the expected answer. There is no misleading or incorrect information. 
\item \textit{4: Accurate:} The answer is mostly accurate, with only minor or negligible inaccuracies. There may be small factual inconsistencies that do not significantly affect the overall meaning or quality of the answer.
\item \textit{3: Neutral: (neither accurate nor inaccurate)}
The answer is somewhat accurate but also contains elements of inaccuracy. It is neither highly accurate nor does it contain substantial errors. 
\item \textit{2: Inaccurate:}
The answer contains multiple factual errors or inaccuracies that detract from its overall quality. While the core meaning might still be understandable, important details are incorrect or misleading.
\item \textit{1: Very Inaccurate:} The answer is largely or completely inaccurate. 
It does not align with the expected or correct information.
\end{itemize}

\textbf{Usefulness:} It evaluates how helpful, relevant, and applicable the answer is for addressing the task or question at hand.
The rating score representing usefulness is as follows: 
\begin{itemize}[label={}]
\item \textit{5: Very Useful:} The answer is highly useful and provides all necessary information in a clear, and concise manner. 
\item \textit{4: Useful:} The answer is useful but may not be exhaustive. It provides relevant information for which question is asked. 
\item \textit{3: Neutral: (neither useful nor not useful)}
The answer is somewhat useful but lacks all information.
\item \textit{2: Slightly Useful:} The answer is minimally useful, offering less information. The overall output does not sufficiently answer the question.
\item \textit{1: Not Useful at All:} The answer is completely unhelpful and irrelevant.
\end{itemize}

\begin{figure*}[h]
    \centering
    \includegraphics [width=12cm]{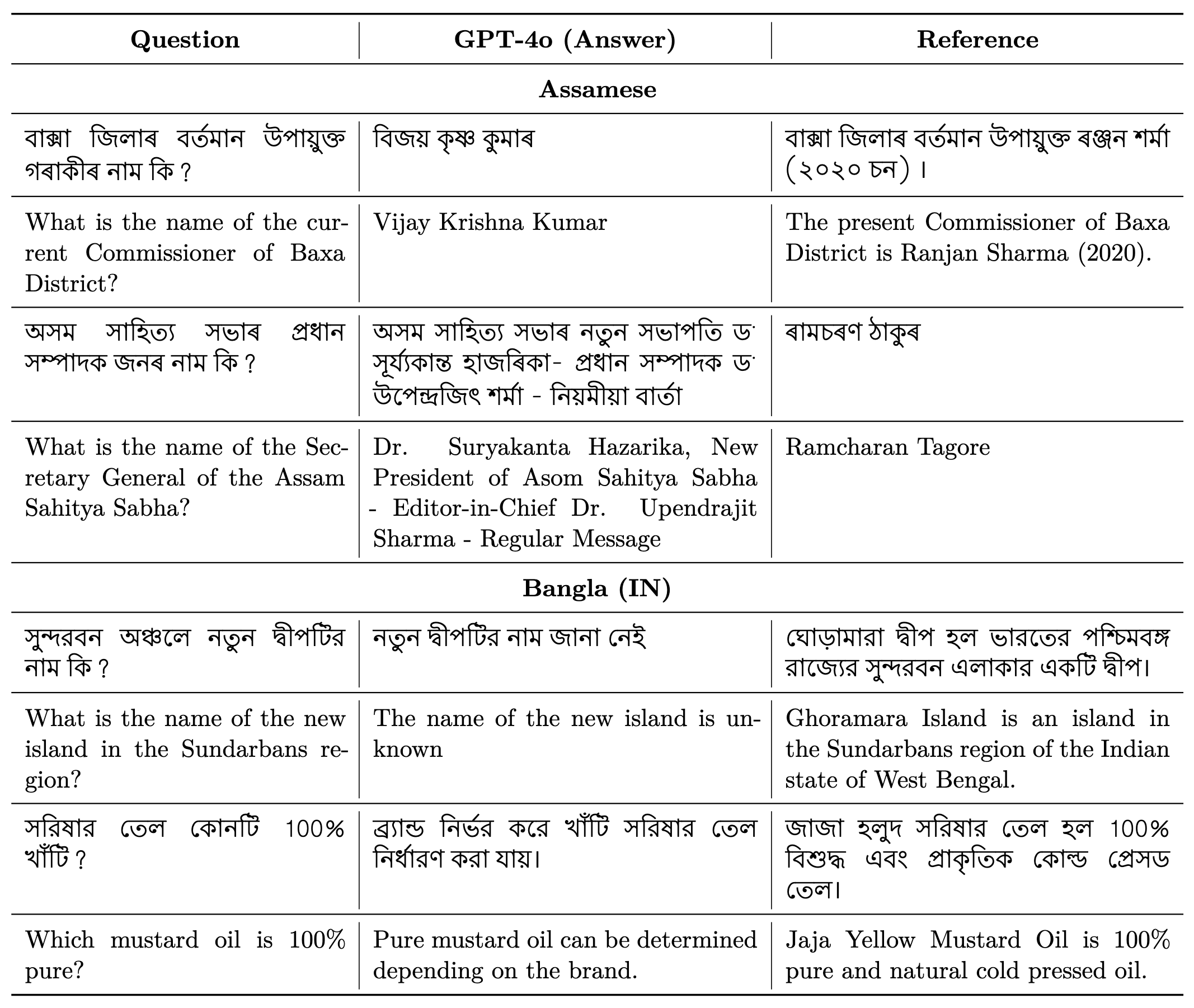}
    \caption{QA pairs with GPT-4o answer and reference for \textit{Assamese} and \textit{Bangla-IN} (with English translation), highlighting potential errors.}
    \label{fig:app_error_analysis_as_bn}
\end{figure*}

\begin{figure*}[t]
    \centering
    \includegraphics [width=12cm]{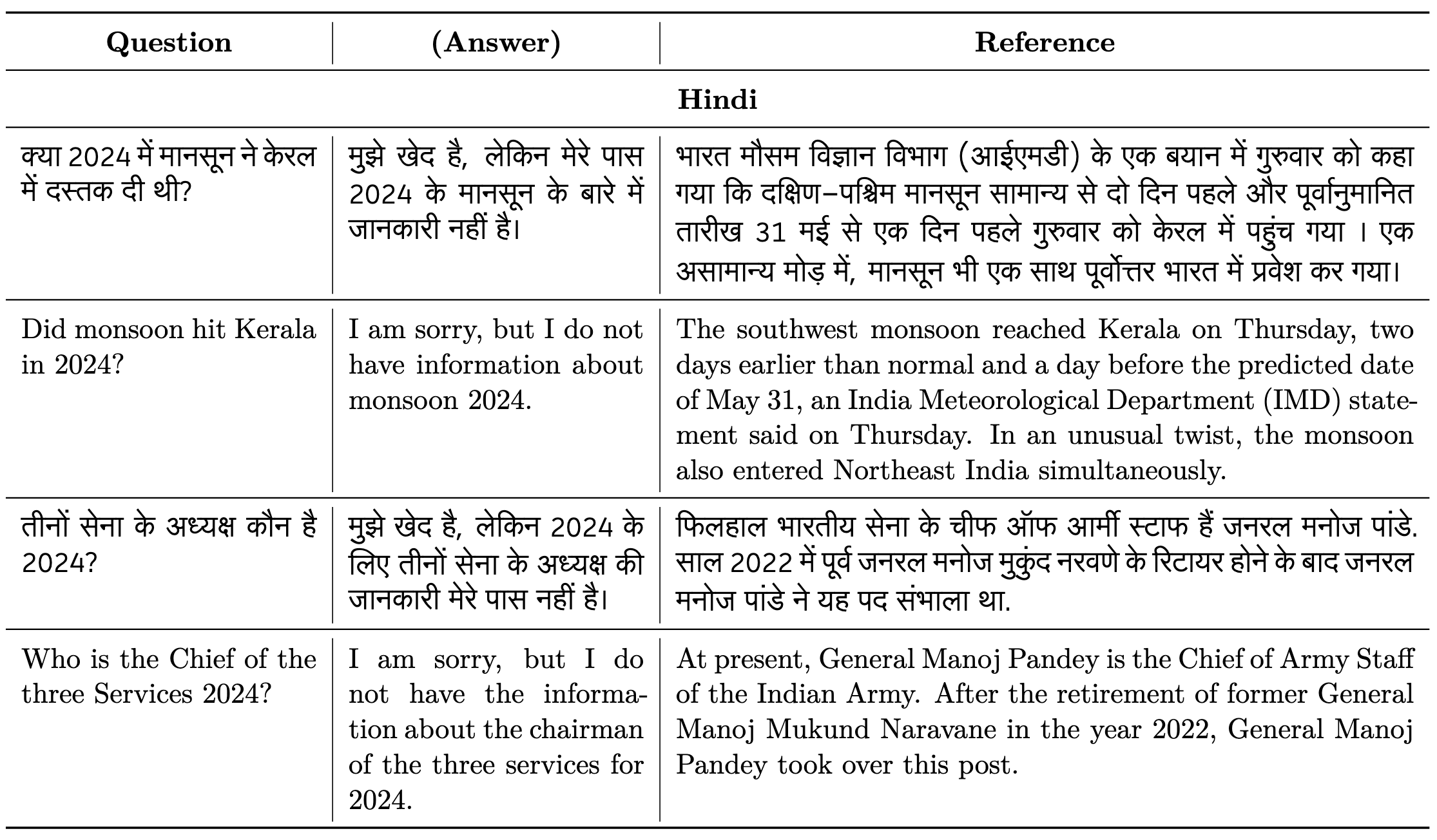}
    \caption{QA pairs with GPT-4o answer and reference for \textit{Hindi} (with English translation), highlighting potential errors.}
    \label{fig:app_error_analysis_hi}
\end{figure*}

\textbf{Human (Subjective) Evaluation:} 
We conducted a human evaluation of the GPT-4o model’s output, focusing on accuracy and usefulness, assessed on a Likert scale (1–5), where higher scores indicate better performance. This evaluation has been done for all languages except Hindi and Nepali and manually checked 100 samples. Following the definitions and instructions provided above, human evaluators scored the answers. Given that this process is time-consuming and costly, we relied on a single annotator for this manual evaluation. While evaluating with multiple annotators would have been ideal, it was not feasible in the current scope of work. 
The results also suggest that GPT-4o is performing well for English and Arabic compared to other languages and comparatively worse for Assamese. This finding is inline with our evaluation using automatic evaluation metrics BLEU and ROUGE. 
In Figure \ref{fig:app_error_analysis_as_bn} and \ref{fig:app_error_analysis_hi}, we report samples of QA pairs for Assamese, Bangla (IN), and Hindi, demonstrating the answer from GPT-4o and reference. Also, it is observed that the GPT-4o answer is short while the reference answer is long. However, it is the opposite in other cases, which impacts the overall performance measures.



\section{Annotation Interface}
\label{sec:app_annotation_interface}

In Figure \ref{fig:app_domain_reliability_interface}, we present a screenshot of the interface designed for domain reliability check, which consisted of a URL of the domain, annotation guidelines, and four different options associated with the four categories we defined for this annotation task. Annotators select one of these labels and submit.

In Figure \ref{fig:app_annotation_step_question_selection} and \ref{fig:app_annotation_step_answer_editing} we provide a screenshot of the interface that demonstrate the steps of question validation, question's relevancy to the location, answer categorization and editing the answer, respectively. The later steps will appear on the interface depending on the classification of the question in the \textit{question validation step}. 



\newsavebox{\tempbox}
\sbox{\tempbox}{\includegraphics[scale=0.265]{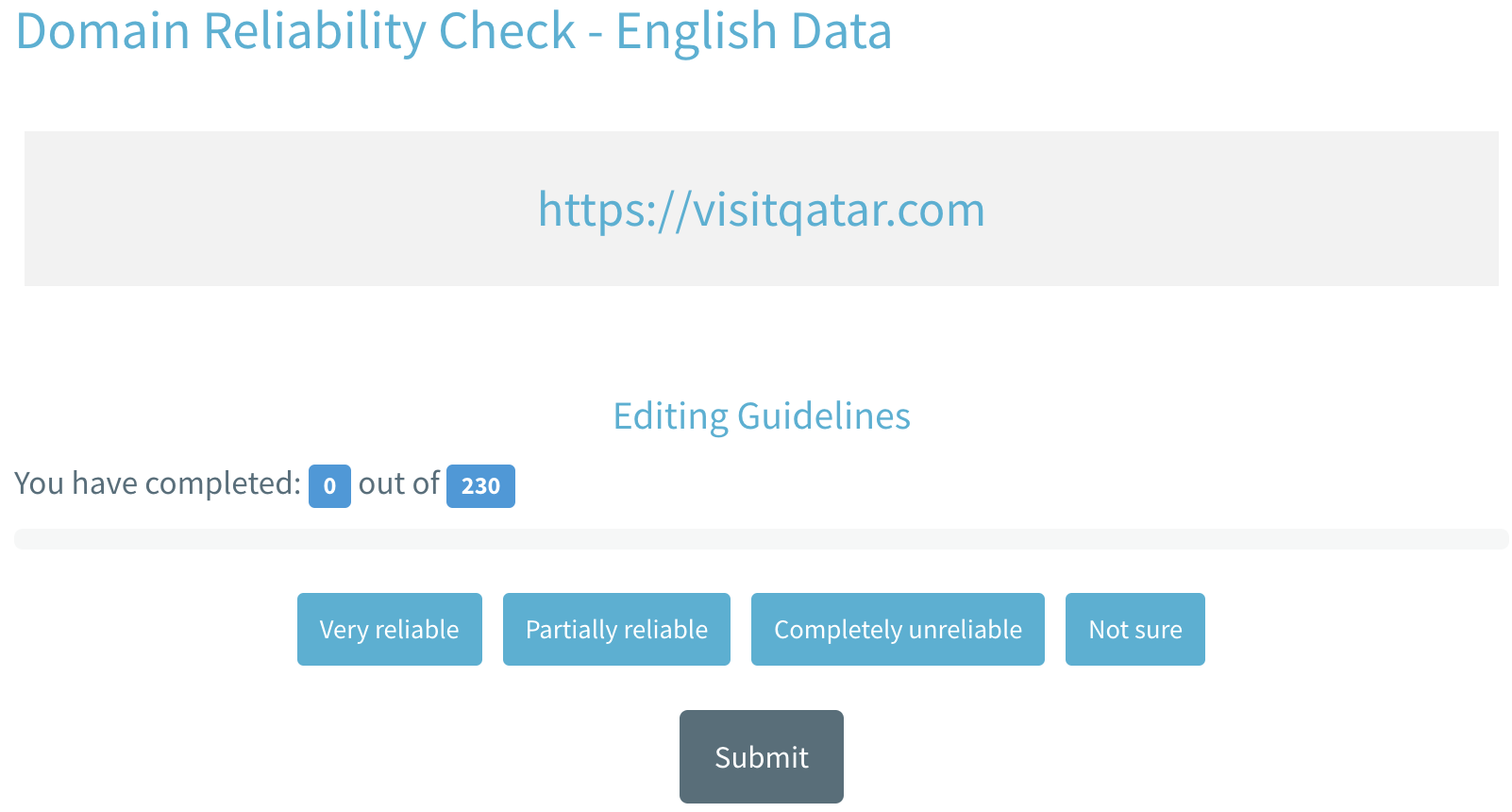}}
\begin{figure*}[t]
    \centering    
    \begin{tikzpicture}
        \node[anchor=south west, inner sep=0] (image) at (0,0) {\usebox{\tempbox}};
        \draw[dotted, thick] (image.south west) rectangle (image.north east);
    \end{tikzpicture}
    \caption{An example of the annotation interface for domain reliability check.}
    \label{fig:app_domain_reliability_interface}
\end{figure*}


\sbox{\tempbox}{\includegraphics[scale=0.8]{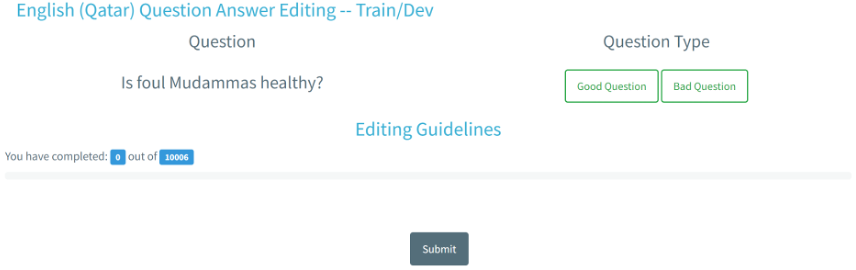}}
\begin{figure*}[t]
    \centering    
    \begin{tikzpicture}
        \node[anchor=south west, inner sep=0] (image) at (0,0) {\usebox{\tempbox}};
        \draw[dotted, thick] (image.south west) rectangle (image.north east);
    \end{tikzpicture}
    \caption{Annotation interface for \textit{Question Validation}.}
        \label{fig:app_annotation_step_question_selection}
\end{figure*}

\sbox{\tempbox}{\includegraphics[scale=0.8]{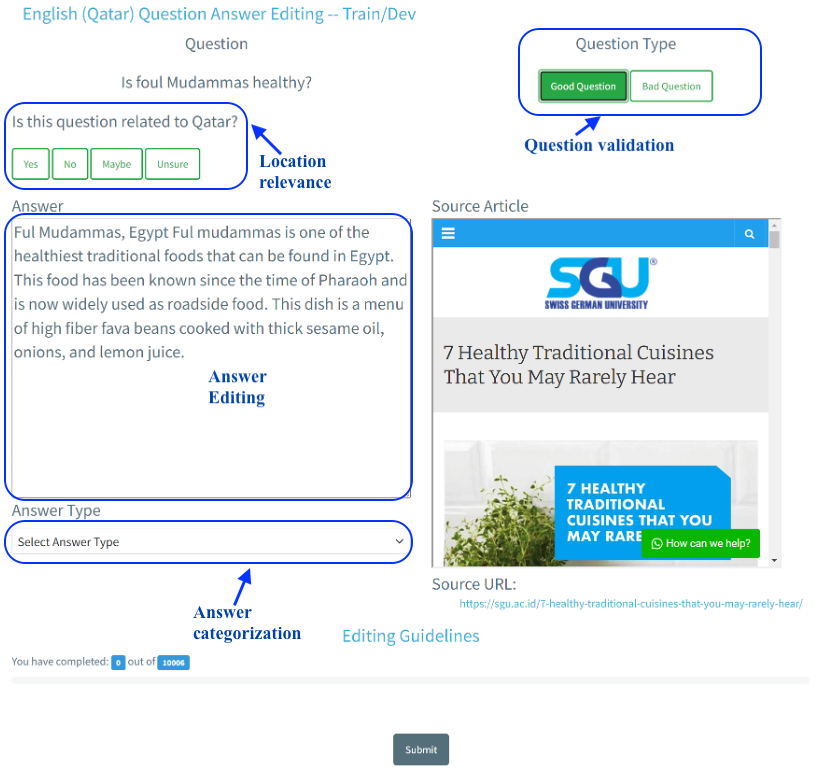}}
\begin{figure*}[t]
    \centering    
    \begin{tikzpicture}
        \node[anchor=south west, inner sep=0] (image) at (0,0) {\usebox{\tempbox}};
        \draw[dotted, thick] (image.south west) rectangle (image.north east);
    \end{tikzpicture}
        \caption{Annotation interface for \textit{question validation}, \textit{location relevance}, \textit{answer editing}, and \textit{answer categorization}.}
        \label{fig:app_annotation_step_answer_editing}
\end{figure*}

\section{Data Release and License}
\label{sec:app_data_release}
The \mnqa{} dataset is publicly available under the Creative Commons Attribution Non Commercial Share Alike 4.0: \url{https://creativecommons.org/licenses/by-nc-sa/4.0/}.

\end{document}